\documentclass[11pt]{article}
\newif\ifanon
\anonfalse
\ifanon\usepackage[review]{acl}\else\usepackage[final]{acl}\fi
\usepackage{times}
\usepackage{latexsym}
\usepackage[T1]{fontenc}
\usepackage[utf8]{inputenc}
\usepackage{microtype}
\usepackage{url}
\usepackage{inconsolata}
\usepackage{graphicx}
\usepackage{booktabs}
\usepackage{amsmath,amssymb}
\usepackage{multirow}
\usepackage{xcolor}
\usepackage{tikz}
\usetikzlibrary{arrows.meta,positioning,calc}
\definecolor{cDag}{HTML}{D55E00}
\definecolor{cDense}{HTML}{6E6E6E}
\definecolor{cLocal}{HTML}{009E73}

\newcommand{\dagf}{DAGFormer}

\newcommand{\valpha}{\boldsymbol{\alpha}}

\title{DAGFormer: Global Topology Prediction for Transformer Cross-Layer Connections}

\ifanon
\author{Anonymous NAACL submission}
\else
\author{Xiaocong Yang\thanks{\ \ Equal contribution.} \quad Yuren Hao\footnotemark[1] \quad ChengXiang Zhai \\
  AI Interpretability @ Illinois \\
  University of Illinois Urbana-Champaign \\
  \texttt{\{xy51, yurenh2, czhai\}@illinois.edu} \\ }
\fi

\begin{document}
\maketitle

\begin{abstract}
We propose \dagf{}, a Transformer language model whose cross-layer connections are planned by a jointly trained topology predictor. Beyond the residual connections in a standard Transformer, \dagf{} adds direct connections from earlier layers to later ones, and the predictor controls the strength of every connection based on input tokens. Unlike existing cross-layer routing architectures, which set each layer's connections separately, \dagf{} plans the routing of the whole model at once. On matched OLMo-2 pairs from 75M to 1B parameters, \dagf{} lowers WikiText-2 perplexity by 17\% to 25\% over the vanilla Transformer, and a 1B \dagf{} trained on 5B tokens outperforms the vanilla model trained on 10B. Against four routing baselines with the same data and budget, it reaches the lowest perplexity at 75M, 300M and 1B. The gain carries over to 13 commonsense, science, reading and math evaluation tasks (2.0 to 2.5 points higher mean accuracy at 300M and 1B) and to instruction finetuning (63\% to 70\% AlpacaEval win rate against the vanilla Transformer, leading every routing baseline), and under structured pruning \dagf{} degrades the least in five of six settings.
\end{abstract}

\section{Introduction}

A Transformer \citep{vaswani2017} composes its layers through residual connections \citep{he2016resnet}: every attention head and MLP reads the same running sum and writes back into it. Mechanistic analyses describe this residual stream as the communication channel between model components \citep{elhage2021}. Its bandwidth is shared by every layer, so which earlier computation a layer can read is fixed by design: a late layer sees an early result only inside a sum to which every intermediate layer has added its own.

\begin{figure*}[t]
\centering
\includegraphics[width=\textwidth]{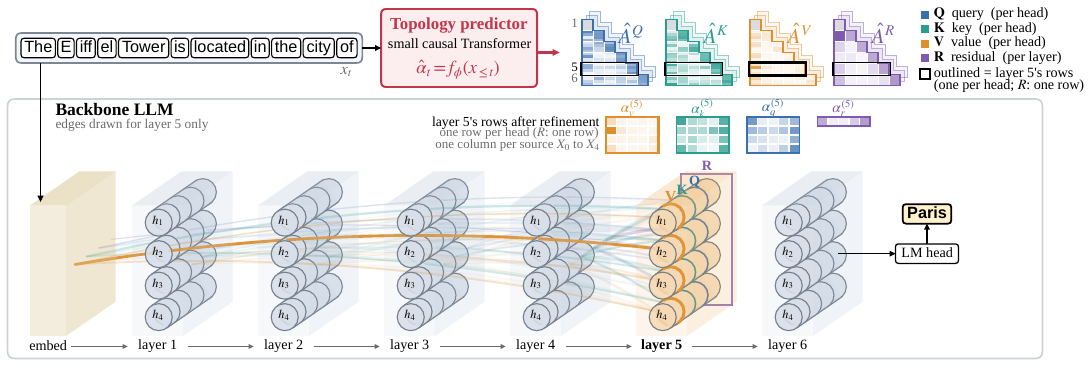}
\caption{\dagf{}. For token $x_t$ the predictor outputs the planned graph as four triangular matrices $\hat{A}^{Q}$, $\hat{A}^{K}$, $\hat{A}^{V}$ and $\hat{A}^{R}$ that collect the planned weights $\hat{\valpha}^{(l)}$ of Eq.~\ref{eq:predictor}: one column per earlier output and, per layer, one row per head for $Q$, $K$ and $V$ (the weights $\hat{\alpha}^{(l)}_{q,h,j}$) and a single row for $R$ (the weights $\hat{\alpha}^{(l)}_{r,j}$). Only the edges into layer~5, its outlined rows, are drawn, as rings on its heads (value in front; key, query and residual behind) with edge width proportional to the magnitude of the entry; every routed layer reads along its own rows in the same way, after the auxiliary local refinement of Eq.~\ref{eq:refine} (not drawn for demonstration). All weights shown are real values from a trained \dagf{} on training data.}
\label{fig:arch}
\end{figure*}

A line of work relaxes this fixed connectivity. Dense connections \citep{huang2017densenet} and depth-weighted averaging with learned static weights \citep{pagliardini2024denseformer} let a layer read earlier layers' outputs directly. Hyper-connections \citep{zhu2024hyperconnections} widen the residual stream into several streams whose mixing weights are learned and, in the dynamic variant, input-dependent, and mHC \citep{mhc2026} constrains those mixing matrices so the scheme trains stably at scale. Multiway dynamic dense connections \citep{muddformer2025} compute input-dependent weights over all earlier layers' outputs, separately for the query, key, value and residual inputs of a block. Each token can then follow its own path through the layers, and these models reach the loss of Transformers trained with substantially more budget \citep{zhu2024hyperconnections,muddformer2025}.

All of the dynamic variants above, however, make the same design choice: routing is \emph{local}. Each layer carries its own router, which reads the hidden state entering that layer and decides what that layer reads and, in hyper-connections and mHC, where its output is written. The cross-layer routing topology is therefore assembled \textit{greedily}, one layer at a time, and no module takes a global view of it. Consequently, cross-layer coordination cannot be planned explicitly from the input, and the topology cannot be inspected as an individual object.

In this paper, we propose \dagf{}, where the global routing topology is predicted with a separate module that reads only the input tokens, before the backbone processes each token. \dagf{} implements this idea on an OLMo-2 backbone \citep{olmo2} with added cross-layer connections on attention heads. Instead of reading the running residual stream, each attention head computes its query, key and value from weighted mixtures of the embedding and all earlier layer outputs. The residual input of each layer is also such a mixture. A small causal encoder predicts all of these mixing weights: it reads the same token ids as the backbone, embedded with its own randomly initialized 256-dimensional table rather than the backbone's embeddings, and is pretrained jointly with the backbone, as shown in Figure~\ref{fig:arch}. Additionally, an auxiliary local refinement, zero-initialized, lets each layer correct the predicted weights of its own heads from its input, so the predictor plans the graph and each layer refines its rows of it.

To study this model architecture, we train matched pairs of the standard OLMo-2 Transformer (vanilla Transformer) and \dagf{} from 75M to 1B parameters on Dolma \citep{soldaini2024dolma}, charging the routing cost in every compute comparison. \dagf{} is better at every size and at equal compute, and a 1B \dagf{} beats the 1B vanilla Transformer trained on twice as many tokens. It also beats four routing baselines trained with the same pipeline at every size but 150M, and the gain carries over to evaluation tasks and instruction finetuning. Ablations show that the shared predictor carries most of the gain, and under structured pruning \dagf{} loses the least of all baselines in five of six settings.

\textbf{Our contributions are as follows:}
\begin{itemize}\setlength\itemsep{0pt}
\item \dagf{}, a Transformer variant with dynamic cross-layer connections on attention heads, where the connections are predicted from the input tokens by a separate topology predictor which is pretrained together with an OLMo-2 backbone from scratch (Section~\ref{sec:method}).
\item A matched scaling study from 75M to 1B with the routing cost charged and with DenseFormer, hyper-connection, mHC and MUDDFormer baselines on the same pipeline, carried through instruction finetuning and instruction-following evaluation (Section~\ref{sec:experiments}).
\item Ablations of the routing source and a structured-pruning study showing that the learned topology makes models degrade more gracefully, without re-planning by the predictor (Sections~\ref{sec:analysis}, \ref{sec:pruning} and~\ref{sec:discussion}).
\end{itemize}

\section{\dagf{}}
\label{sec:method}

\dagf{} pairs a Transformer backbone with a small \emph{topology predictor} that reads the input tokens and predicts a weighted directed acyclic graph assigning every attention head and every layer its sources among the earlier outputs; the backbone computes along this graph (Figure~\ref{fig:arch}). The predictor, a causal encoder over the tokens, plans the graph for the whole depth at once; an auxiliary local refinement, a tiny network at each layer, refines that layer's planned rows from the layer's own input.

\subsection{Setup}
\label{sec:setup}

Let $X_0(t)\in\mathbb{R}^D$ be the embedding of token $t$ and $X_l(t)$ the output of layer $l$ in an $L$-layer decoder of width $D$ with $H$ attention heads; we omit $t$ when it is clear from context. A vanilla OLMo-2 Transformer layer \citep{olmo2} reads $X_{l-1}$ $3H{+}1$ times: head $h$ forms $q_h = W_q^h X_{l-1}$, $k_h = W_k^h X_{l-1}$ and $v_h = W_v^h X_{l-1}$, and the residual $R_l = X_{l-1}$ carries the stream forward, with the update
\begin{align*}
X_l &= R_l + a_l + m_l,\\
a_l &= \mathrm{Norm}\big(\mathrm{Attn}_l(q,k,v)\big),\\
m_l &= \mathrm{Norm}\big(\mathrm{MLP}_l(R_l + a_l)\big).
\end{align*}
All $3H{+}1$ reads see the same vector, and an earlier layer's output reaches them only inside the sum $X_{l-1}$. \dagf{} instead gives every read of a routed layer ($l\ge2$) its own weight on each earlier output $X_0,\dots,X_{l-1}$; the $(3H{+}1)\,l$ weights of layer $l$, written $\valpha^{(l)}(t)$, together over all routed layers make up the graph of token $t$, and they fill the four triangular matrices of Figure~\ref{fig:arch}, one row per read and one column per source.

\subsection{Topology predictor}
\label{sec:predictor}

The input sequence first goes into a small network, called the topology predictor, that generates the connection graph for the backbone LLM. The predictor embeds the tokens with its own token and position embeddings of width 256, encodes them with two causal Transformer layers with four heads, passes each output vector $z$ through a shared trunk $u=\mathrm{GELU}(W_u\,\mathrm{LN}(z))\in\mathbb{R}^{512}$, and ends in one linear head per routed layer, which produces that layer's adjacency matrix, with rows for the query, key and value reads of every head and for the residual (Figure~\ref{fig:arch}, top):
\begin{equation}
\hat{\valpha}^{(l)} = W^{(l)} u + b^{(l)},
\label{eq:predictor}
\end{equation}
where $W^{(l)}\in\mathbb{R}^{(3H+1)l\times 512}$ and $b^{(l)}\in\mathbb{R}^{(3H+1)l}$ are the head's weight and bias; the $(3H{+}1)\,l$ outputs are reshaped into the matrix. Its entries are signed mixing coefficients: unconstrained real values that, unlike attention probabilities, are not softmax-normalized, so a read can amplify a source or subtract one source from another. $W^{(l)}$ starts at zero and every row of $b^{(l)}$ at the one-hot vector that selects $X_{l-1}$. The graph thus depends on the input and changes as the sequence grows during generation.

As an auxiliary component, each layer refines its planned matrix with a local term computed by a tiny network from the layer input $X_{l-1}$:
\begin{equation}
\valpha^{(l)} = \hat{\valpha}^{(l)} + W_2^{(l)}\,\mathrm{GELU}\big(W_1^{(l)}\,\mathrm{LN}(X_{l-1})\big),
\label{eq:refine}
\end{equation}
where $W_1^{(l)}\in\mathbb{R}^{128\times D}$ and $W_2^{(l)}\in\mathbb{R}^{(3H+1)l\times 128}$, with $W_2^{(l)}$ initialized to zero. The refinement lets a layer correct the plan with what the backbone has computed so far, which the predictor, reading only the tokens, cannot see. It acts at the granularity of the plan, one correction per head and read, and it corrects the predicted weights rather than producing them.

Since the encoder is causal, the graph of a token never depends on later tokens, and \dagf{} generates left to right like any decoder. All layers are planned from the same encoding of the sequence, because the trunk is shared; a deeper backbone only adds linear heads. The planned term never sees the backbone and is fixed before the first layer runs, which is what lets Section~\ref{sec:analysis} read the graph, average it over inputs, freeze it, and swap it between inputs with the backbone unchanged.

\begin{table*}[t]
\centering\footnotesize
\setlength{\tabcolsep}{3.6pt}
\begin{tabular}{l rrrrr r r}
\toprule
& \multicolumn{5}{c}{baselines (WikiText-2 NLL $\downarrow$)} & & \\
\cmidrule(lr){2-6}
size & vanilla & DenseFormer & hyper-conn. & mHC & MUDDFormer & \dagf{} & \textbf{gap to vanilla} \\
\midrule
75M & 4.997 & 5.027 & 4.964 & \underline{4.913} & 4.974 & \textbf{4.711} & \textbf{+0.29} \\
150M & 4.268 & 4.246 & 4.220 & 4.209 & \textbf{4.041} & \underline{4.085} & \textbf{+0.18} \\
300M & 3.808 & 3.745 & 3.692 & 3.652 & \underline{3.649} & \textbf{3.544} & \textbf{+0.26} \\
1B, 5B tokens & 3.569 & 3.513 & 3.492 & 3.438 & \underline{3.423} & \textbf{3.307} & \textbf{+0.26} \\
1B, 10B tokens & 3.446 & -- & -- & -- & -- & \textbf{3.219} & \textbf{+0.23} \\
\midrule
& \multicolumn{6}{c}{prefill time per token relative to vanilla $\downarrow$} & \\
\cmidrule(lr){2-7}
75M & 1.00 & 1.01 & 1.99 & 1.75 & 1.40 & 1.49 & \\
150M & 1.00 & 1.02 & 2.11 & 1.78 & 1.48 & 1.61 & \\
300M & 1.00 & 1.02 & 2.14 & 1.74 & 1.54 & 1.89 & \\
1B & 1.00 & 1.01 & 1.90 & 1.54 & 1.36 & 1.81 & \\
\bottomrule

\end{tabular}
\caption{Held-out WikiText-2 NLL (nats per token); every model of a row shares recipe, data, token budget and seed. Gap to vanilla: vanilla $-$ \dagf{}. Bottom: measured prefill time per token relative to the vanilla Transformer (a forward pass over 8 sequences of 1024 tokens); training cost in Appendix~\ref{app:flops}. Bold: best per row; underlined: second best (not marked at 1B, 10B tokens, with two models). --: not run (the baselines are trained at 5B tokens only due to computing budget limit).}
\label{tab:main}
\end{table*}

\subsection{Routed layers}
\label{sec:routed}

The backbone LLM then runs along this graph. In routed layer $l$, each head forms its query, key and value from weighted sums of the earlier outputs, and so does the residual:
\begin{equation}
\begin{aligned}
q_h &= \sum_{j<l}\alpha^{(l)}_{q,h,j}\,W_q^h X_j, & k_h &= \sum_{j<l}\alpha^{(l)}_{k,h,j}\,W_k^h X_j,\\
v_h &= \sum_{j<l}\alpha^{(l)}_{v,h,j}\,W_v^h X_j, & R_l &= \sum_{j<l}\alpha^{(l)}_{r,j}\,X_j,
\end{aligned}
\label{eq:mix}
\end{equation}
where $\alpha^{(l)}_{q,h,j}$ is the entry of $\valpha^{(l)}$ in the row of head $h$'s query read and the column of source $X_j$, and likewise for $k$, $v$ and $r$. Other modules, attention with Q/K normalization on $q_h$ and $k_h$, the MLP on $R_l+a_l$, and the layer output $X_l=R_l+a_l+m_l$, remain unchanged from the standard Transformer; at 300M and 1B, \dagf{} also applies an RMSNorm to the mixed values $v_h$ (V normalization; Table~\ref{tab:hparams}). In practice, each layer runs its fused query, key and value projection once over the stacked sources and then takes the weighted sums of Eq.~\ref{eq:mix} in head space. Layer 1 has the embedding as its only source and is not routed. At initialization every row of $\valpha^{(l)}$ selects $X_{l-1}$, so Eq.~\ref{eq:mix} reduces to the vanilla layer and \dagf{} starts as exactly the vanilla OLMo-2 Transformer, up to the V normalization where enabled; predictor and backbone are then trained jointly from scratch on the language-modeling loss.

\section{Experiments}
\label{sec:experiments}

\subsection{Data}
\label{sec:data}

All pretraining uses Dolma v1.7 \citep{soldaini2024dolma}. Three tokenizations were used (a 12B-token slice for the 75M to 600M pairs, a local rebuild of it, and a 21B-token one for the 300M and 1B pairs), so every final checkpoint is re-evaluated on the same held-out sets: WikiText-2 \citep{merity2016wikitext}, a held-out slice of MathInstruct \citep{yue2023mammoth}, GSM8K solutions \citep{cobbe2021gsm8k}, and a source-balanced Dolma set drawn from files no training corpus consumed (Appendix~\ref{app:cache}). WikiText-2 is the primary metric; the balanced set gives the same orderings with gaps 10\% to 25\% smaller. For instruction finetuning we use two datasets: Alpaca \citep{taori2023alpaca} in its cleaned release plus Dolly \citep{conover2023dolly}, about 67K single-turn instructions, and the smol-smoltalk subset of SmolTalk \citep{allal2025smollm2}, about 460K conversations, the mix used to finetune SmolLM2 at 135M and 360M parameters. The held-out instruction data of Table~\ref{tab:instruct} are 1\% of Alpaca plus Dolly and the test split of smol-smoltalk.

\subsection{Setup and baselines}
\label{sec:expsetup}

\paragraph{Matched pairs.} Every comparison is between a vanilla OLMo-2 Transformer and a \dagf{} with the identical backbone, tokenizer, data order, token budget and optimizer. Sizes are 75M (6 layers, $D{=}512$, 8 heads), 150M (8, 768, 12), 300M (12, 1024, 16) and 1B (16, 2048, 16), counting backbone parameters with tied embeddings; routing adds 29M to 37M parameters, 25.7M of them the predictor's token embedding. Sequences are 1024 tokens, 524K tokens per step (Table~\ref{tab:hparams}). The 75M, 150M and 300M runs train for about 21 tokens per parameter, close to the compute-optimal ratio of about 20 implied by \citet{hoffmann2022chinchilla}; the 1B pair trains for 5B tokens and is continued to 10B with a re-warmed schedule, giving matched pairs at both budgets. Seed 42 fixes initialization and data order; the 300M pair is repeated with a second seed.

\begin{table*}[t]
\centering\footnotesize
\begin{tabular*}{\textwidth}{@{\extracolsep{\fill}}cc cccc @{\hspace{6pt}}|c c}
\toprule
 & & \multicolumn{4}{c}{mean accuracy (\%) per kind of task $\uparrow$} & & \\
\cmidrule(lr){3-6}
size & model & commonsense (5) & science (4) & reading (2) & math (2) & {\fontsize{9pt}{10pt}\selectfont \textbf{Mean}} & mean rank $\downarrow$ \\
\midrule
300M & vanilla & \underline{39.8} & 36.5 & 38.1 & 11.7 & {\fontsize{9pt}{10pt}\selectfont 34.2} & 4.15 \\
 & DenseFormer & 38.9 & 36.5 & 40.8 & 12.7 & {\fontsize{9pt}{10pt}\selectfont 34.4} & 4.23 \\
 & hyper-conn. & 38.5 & 38.5 & 38.3 & 12.3 & {\fontsize{9pt}{10pt}\selectfont 34.4} & 3.69 \\
 & mHC & 39.2 & \underline{38.8} & \underline{41.8} & \textbf{12.9} & {\fontsize{9pt}{10pt}\selectfont 35.4} & 2.54 \\
 & MUDDFormer & 39.7 & 38.6 & 40.9 & 11.5 & {\fontsize{9pt}{10pt}\selectfont 35.2} & 3.08 \\
 & \dagf{} & \textbf{39.9} & \textbf{39.8} & \textbf{43.5} & \underline{12.8} & {\fontsize{9pt}{10pt}\selectfont \textbf{36.2}} & \textbf{2.38} \\
\midrule
1B & vanilla & 40.0 & 38.9 & 42.1 & 12.1 & {\fontsize{9pt}{10pt}\selectfont 35.7} & 4.54 \\
(5B) & DenseFormer & 40.2 & 39.5 & 43.0 & 11.4 & {\fontsize{9pt}{10pt}\selectfont 36.0} & 4.38 \\
 & hyper-conn. & 40.7 & 40.7 & 41.6 & 12.0 & {\fontsize{9pt}{10pt}\selectfont 36.4} & 3.77 \\
 & mHC & \underline{41.1} & 42.2 & \underline{43.8} & \textbf{12.5} & {\fontsize{9pt}{10pt}\selectfont 37.4} & 2.62 \\
 & MUDDFormer & 40.7 & \underline{42.4} & 38.3 & 12.2 & {\fontsize{9pt}{10pt}\selectfont 36.5} & 3.15 \\
 & \dagf{} & \textbf{41.8} & \textbf{43.0} & \textbf{44.7} & \underline{12.4} & {\fontsize{9pt}{10pt}\selectfont \textbf{38.1}} & \textbf{2.08} \\
\midrule
1B & vanilla & 40.5 & 41.0 & 43.0 & 11.5 & {\fontsize{9pt}{10pt}\selectfont 36.6} & -- \\
(10B) & \dagf{} & \textbf{42.4} & \textbf{44.1} & \textbf{47.0} & \textbf{12.8} & {\fontsize{9pt}{10pt}\selectfont \textbf{39.1}} & -- \\
\bottomrule
\end{tabular*}

\caption{Base models on the evaluation tasks of Section~\ref{sec:expsetup}: mean accuracy (\%) over the commonsense (HellaSwag, PIQA, WinoGrande, CommonsenseQA, Social IQa), science (ARC-Easy/Challenge, OpenBookQA, SciQ), reading (BoolQ, LAMBADA) and math tasks (MathQA, GSM8K; close to chance at these sizes), and over all 13 (Mean). Mean rank: rank within the block, averaged over the 13 tasks (1 = best). Bold: best per block; underlined: second best per kind of task (neither marked nor ranked in the two-model block). 1B (5B), 1B (10B): pretrained on 5B and 10B tokens. Per-task accuracies in Table~\ref{tab:downstream_tasks}.}
\label{tab:downstream}
\end{table*}

\paragraph{Compute.} Projecting the $l$ sources of layer $l$ and then mixing them, as in training, adds $\sum_{l\ge2} (l{-}1)\cdot 3D^2$ multiply-accumulates over the vanilla Transformer's $\sum_l (4D^2 + 3DI)$, plus the mixing sums, the predictor and the local refinements: 1.18$\times$ the vanilla Transformer's FLOPs per token at 75M, rising to 2.13$\times$ at 1B; mixing per head before projecting gives the same result at 1.02 to 1.04$\times$ but is memory-bound (Appendix~\ref{app:orders}). Compiled and in the faster order (Table~\ref{tab:main}, bottom), a \dagf{} token takes 1.5 to 1.9$\times$ the vanilla Transformer's prefill time, the predictor 1 to 3\% of it: more than MUDDFormer (1.4 to 1.5$\times$) and, from 300M, mHC (1.5 to 1.8$\times$), less than hyper-connections (1.9 to 2.1$\times$); DenseFormer adds nothing measurable. MUDDFormer and the hyper-connection variants add few FLOPs (MUDDFormer reports 0.4\%; \citealp{muddformer2025}), so most of their cost lies outside the FLOP count; none of the implementations uses hand-written kernels, with which mHC reports a 6.7\% training-time overhead at four streams \citep{mhc2026}. Training costs are in Appendix~\ref{app:flops}; compute-matched comparisons charge the analytic training FLOPs of the training order.

\paragraph{Baselines.} Besides the matched vanilla Transformer, we train four cross-layer baselines with the recipe, data, tokens and seed of the pair of the same size: DenseFormer \citep{pagliardini2024denseformer}, whose cross-layer weights are static; dynamic hyper-connections with four streams \citep{zhu2024hyperconnections}; manifold-constrained hyper-connections (mHC; \citealp{mhc2026}); and MUDDFormer \citep{muddformer2025}, whose per-layer routers read the hidden state. Implementation details, parameter-matched vanilla Transformers and static hyper-connections are in Appendix~\ref{app:posthoc}. We do not compare with mixture-of-experts or mixture-of-depths models, which route computation rather than connectivity.

\begin{table}[t]
\centering\footnotesize
\setlength{\tabcolsep}{3pt}
\begin{tabular}{ll rc rc}
\toprule
 & & \multicolumn{2}{c}{+Alpaca} & \multicolumn{2}{c}{+SmolTalk} \\
\cmidrule(lr){3-4}\cmidrule(lr){5-6}
size & model & NLL $\downarrow$ & win $\uparrow$ & NLL $\downarrow$ & win $\uparrow$ \\
\midrule
300M & vanilla & 2.392 & -- & 1.760 & -- \\
 & DenseFormer & 2.342 & 0.55 & 1.722 & 0.57 \\
 & hyper-conn. & 2.298 & 0.59 & 1.684 & 0.58 \\
 & mHC & 2.277 & 0.58 & 1.664 & 0.62 \\
 & MUDDFormer & 2.256 & 0.63 & 1.639 & 0.64 \\
 & \dagf{} & \textbf{2.203} & \textbf{0.65} & \textbf{1.595} & \textbf{0.70} \\
\midrule
1B (5B) & vanilla & 2.208 & -- & 1.728 & -- \\
 & DenseFormer & 2.186 & 0.46 & 1.706 & 0.50 \\
 & hyper-conn. & 2.150 & 0.51 & 1.673 & 0.56 \\
 & mHC & 2.110 & 0.55 & 1.633 & 0.61 \\
 & MUDDFormer & 2.104 & 0.57 & 1.617 & 0.56 \\
 & \dagf{} & \textbf{2.042} & \textbf{0.63} & \textbf{1.558} & \textbf{0.66} \\
\midrule
1B (10B) & vanilla & 2.120 & -- & 1.669 & -- \\
 & \dagf{} & \textbf{1.974} & \textbf{0.66} & \textbf{1.516} & \textbf{0.70} \\
\bottomrule
\end{tabular}

\caption{Instruction following after finetuning. NLL: held-out negative log-likelihood on the instruction data. Win: AlpacaEval \citep{alpaca_eval} win rate against the vanilla Transformer of the same size and recipe, both answer orders judged by \texttt{gpt-5.6-sol}, ties half, 249 of the 805 prompts. 1B (5B), 1B (10B): the 1B models pretrained on 5B and 10B tokens.}
\label{tab:instruct}
\end{table}

\paragraph{Evaluation tasks.} Base models are evaluated with the lm-evaluation-harness \citep{gao2024lmeval} on twelve tasks (ARC-Easy/Challenge, HellaSwag, PIQA, WinoGrande, OpenBookQA, SciQ, BoolQ, LAMBADA, CommonsenseQA, Social IQa, MathQA; \citealp{clark2018arc,zellers2019hellaswag,bisk2020piqa,sakaguchi2020winogrande,mihaylov2018openbookqa,welbl2017sciq,clark2019boolq,paperno2016lambada,talmor2019commonsenseqa,sap2019socialiqa,amini2019mathqa}) and on GSM8K (3-shot exact match with flexible numeric extraction, on the first 200 test problems). The models finetuned on each instruction dataset (Section~\ref{sec:data}; hyperparameters in Appendix~\ref{app:hparams}) are compared by negative log-likelihood on held-out instruction data and by the AlpacaEval \citep{alpaca_eval} win rate against the vanilla Transformer of the same size and recipe, judged by \texttt{gpt-5.6-sol}; an MT-Bench comparison \citep{zheng2023judge} is in Appendix~\ref{app:downstream}.

\begin{figure*}[t]
\centering
\includegraphics[width=\textwidth]{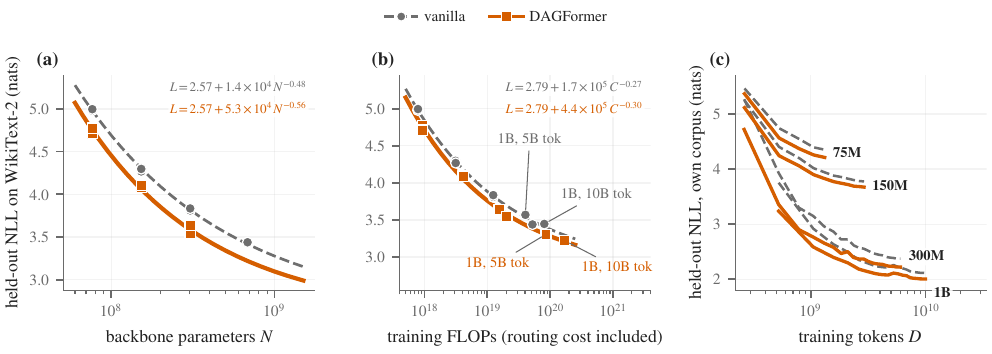}
\caption{Scaling of \dagf{} (solid) against the matched vanilla OLMo-2 Transformer (dashed). (a) Final WikiText-2 loss against backbone parameters, with power laws whose irreducible loss is shared by the two families (coefficients in the panel); the 1B runs, trained on fewer tokens per parameter, appear only in (b). (b) Loss against training FLOPs with the routing cost included, fitted likewise: the 1B \dagf{} at 5B tokens is below the 1B vanilla Transformer at 10B tokens at about the same compute. (c) Held-out loss along training for the four matched pairs: the gap opens early and persists.}
\label{fig:scaling}
\end{figure*}

\subsection{Main results}
\label{sec:main}

\dagf{} has the lowest WikiText-2 loss of all baselines at every size but 150M (Table~\ref{tab:main}). It is 0.18 to 0.29 nats below the vanilla Transformer at every size (17\% to 25\% lower perplexity) and 0.1 to 0.2 nats below the strongest routing baseline at 75M, 300M and 1B; at 150M MUDDFormer is 0.04 nats lower (Section~\ref{sec:discussion}). The gap to the vanilla Transformer opens early in training (Figure~\ref{fig:scaling}c) and holds on the other held-out sets, a second corpus and seed, and across learning rates (Appendix~\ref{app:pairs}, Table~\ref{tab:lrsweep}). On the 13 downstream evaluation tasks, \dagf{} has the highest mean accuracy of all models at both sizes we tested (Table~\ref{tab:downstream}): 2.0 to 2.5 points above the vanilla Transformer and 0.7 to 0.8 points above the strongest routing baseline. It also leads on the commonsense, science and reading tasks and has the best mean rank at both sizes; on math, where MathQA and GSM8K are close to chance, mHC is 0.1 points ahead. At 1B with 10B tokens, \dagf{} is ahead of the vanilla Transformer on all 13 tasks.

We further evaluate the instruction-finetuned models on held-out instruction data and on commonly used instruction-following benchmarks (Table~\ref{tab:instruct}). \dagf{} has the lowest negative log-likelihood on held-out instruction data of all models at every size and under both recipes, 0.15 to 0.19 nats below the vanilla Transformer and 0.04 to 0.06 below the strongest routing baseline. On AlpacaEval it wins 0.63 to 0.70 of the comparisons against the vanilla Transformer, above every routing baseline (0.46 to 0.64). On the 80 MT-Bench first turns, judged pairwise by a second LLM, \dagf{} wins 0.56 to 0.69 of the comparisons against the vanilla Transformer, and the 95\% interval excludes 0.5 in seven of ten (Table~\ref{tab:chat_full}).

\section{Analysis}
\label{sec:analysis}

\subsection{Scaling with compute and parameters}
\label{sec:scaling}

\begin{table}[t]
\centering\scriptsize
\setlength{\tabcolsep}{2.4pt}
\begin{tabular}{llrrrrr}
\toprule
 & & & \multicolumn{4}{c}{held-out NLL $\downarrow$} \\
\cmidrule(lr){4-7}
size & model & params & WikiText & Math & GSM8K & balanced \\
\midrule
75M & vanilla & 77M & 4.997 & 4.271 & 4.242 & 5.167 \\
 & vanilla, hidden 656 & 107M & 4.769 & 4.057 & 4.052 & 4.961 \\
 & vanilla, 13 layers & 106M & 4.871 & 4.215 & 4.174 & 5.094 \\
 & \dagf{} & 106M & \textbf{4.711} & \textbf{3.984} & \textbf{3.973} & \textbf{4.937} \\
\addlinespace[2pt]
150M & vanilla & 153M & 4.268 & 3.599 & 3.608 & 4.471 \\
 & vanilla, hidden 864 & 182M & 4.199 & 3.537 & 3.599 & 4.409 \\
 & \dagf{} & 183M & \textbf{4.085} & \textbf{3.383} & \textbf{3.402} & \textbf{4.305} \\
\bottomrule
\end{tabular}
\caption{Capacity controls: vanilla Transformers given \dagf{}'s routing parameters as extra width or, at 75M, as extra depth, trained with the pair's recipe, data, tokens and seed (12B' corpus). Params: total, including the predictor and refinement. Math: MathInstruct; balanced: the balanced Dolma set. Bold: best per size and column.}
\label{tab:capacity}
\end{table}

For each family we fit $L=E+AN^{-\alpha}$ to the final WikiText-2 loss against backbone parameters $N$ and $L=E+KC^{-\gamma}$ against training compute $C$, routing cost included, with the irreducible loss $E$ shared by the two families (Figure~\ref{fig:scaling}). We share $E$ because it estimates the entropy of the evaluation text, which does not depend on the architecture. At equal training compute the margin is larger at 300M and 1B than at the smaller sizes: on the vanilla compute fit, the vanilla Transformer needs 1.2 to 1.4$\times$ \dagf{}'s training compute to reach its loss at 75M and 150M, and 1.7 to 2.0$\times$ at 300M and 1B. \dagf{}'s fitted exponents are also larger (0.56 against 0.48 for parameters, 0.30 against 0.27 for compute), so on the fit the compute multiplier keeps growing (1.9$\times$ at $10^{20}$ FLOPs and 2.3$\times$ at $10^{21}$), although in each fit the two exponents differ by about one standard error and nothing beyond our largest run is tested. Without any fit, the 1B \dagf{} trained on 5B tokens beats the 1B vanilla Transformer trained on 10B tokens at about the same training compute, on WikiText-2, MathInstruct and GSM8K.

\begin{table*}[t]
\centering\small
\setlength{\tabcolsep}{2pt}
\begin{tabular*}{\textwidth}{@{\extracolsep{\fill}}l rrr rrr rrr}
\toprule
 & \multicolumn{3}{c}{75M} & \multicolumn{3}{c}{150M} & \multicolumn{3}{c}{300M} \\
\cmidrule(lr){2-4}\cmidrule(lr){5-7}\cmidrule(lr){8-10}
 & NLL $\downarrow$ & \multicolumn{2}{c}{damage (\%) $\downarrow$} & NLL $\downarrow$ & \multicolumn{2}{c}{damage (\%) $\downarrow$} & NLL $\downarrow$ & \multicolumn{2}{c}{damage (\%) $\downarrow$} \\
model & 0\% & 50\% & 70\% & 0\% & 50\% & 70\% & 0\% & 50\% & 70\% \\
\midrule
vanilla & 2.387 & 9.0 & \textbf{17.9} & 1.854 & 9.9 & 21.2 & 1.676 & 7.4 & 17.8 \\
DenseFormer & 2.383 & 9.5 & 18.7 & 1.830 & 9.5 & 20.7 & 1.641 & 7.9 & 18.5 \\
hyper-conn. & 2.306 & 9.5 & 19.5 & 1.804 & 9.7 & 21.4 & 1.604 & 7.7 & 17.8 \\
mHC & 2.328 & 9.0 & 20.1 & 1.813 & 9.5 & 21.0 & 1.585 & 8.4 & 18.4 \\
MUDDFormer & 2.362 & 10.6 & 22.3 & \textbf{1.692} & 12.8 & 28.6 & 1.558 & 9.3 & 22.0 \\
\dagf{} & \textbf{2.185} & \textbf{7.6} & 22.2 & 1.724 & \textbf{9.2} & \textbf{20.2} & \textbf{1.507} & \textbf{6.9} & \textbf{16.5} \\
\bottomrule
\end{tabular*}

\caption{Pruning during domain finetuning on MathInstruct, every model of Table~\ref{tab:main} from its own checkpoint (75M and 150M: 12B' corpus; 300M: 21B): held-out NLL of the unpruned finetune (0\%) and the damage (loss increase relative to it) at 50\% and 70\% backbone sparsity. Bold: best per column. All sparsities and the frozen-router controls: Table~\ref{tab:pruning_full}.}
\label{tab:pruning}
\end{table*}

Extra parameters explain most of the gain at the smallest size and less than half at the larger one. Vanilla Transformers given \dagf{}'s routing parameters as extra width close 70 to 89\% of the gap at 75M but only 4 to 42\% at 150M, extra depth closes less, and \dagf{} stays ahead of every control on every held-out set (Table~\ref{tab:capacity}).

\subsection{Where the gain comes from}
\label{sec:source}

The predictor carries most of the gain, and the refinement completes it. The routing-source ablation varies only where $\valpha$ comes from in one 75M backbone (Table~\ref{tab:locality}, Appendix~\ref{app:locality}). The predictor alone, which never sees a hidden state, recovers about 90\% of \dagf{}'s gain over the vanilla Transformer, and adding the refinement gives the best result on every held-out set. Sharing matters: independent per-layer predictors reading the same tokens recover under 80\%, with 7M more parameters.

The refinement is cheap, under 0.5\% of \dagf{}'s forward multiply-accumulates, and unlike the per-layer routers of hyper-connections and MUDDFormer, which produce a layer's weights on their own, it corrects the predictor's plan per attention head. \dagf{}'s cost lies instead in projecting every earlier output through each layer's query, key and value weights before mixing (Appendix~\ref{app:flops}), where MUDDFormer mixes the hidden states first and projects once.

\section{Structured pruning}
\label{sec:pruning}

If the benefit is structural, heads that read many earlier layers should lose less when units are removed, without the routers re-planning. We finetune every model of Table~\ref{tab:main} up to 300M from its own checkpoint on MathInstruct \citep{yue2023mammoth} and, during finetuning, permanently remove attention heads and MLP channels (two of CoFi's unit types; \citealp{xia2022cofi}) by first-order Taylor importance on unit gates \citep{michel2019heads,molchanov2019taylor} on a gradual cubic schedule \citep{zhu2017prune}, followed by recovery steps. Routers are never pruned, and we report backbone sparsity.

\dagf{} loses the least of all models in five of the six settings (Table~\ref{tab:pruning}). With half the heads and channels removed, it degrades less than the vanilla Transformer and every routing baseline at every size, and remains about as good as or better than the unpruned vanilla Transformer. With 70\% removed, it degrades the least at 150M and 300M; at 75M the vanilla Transformer, DenseFormer and both hyper-connection variants lose less, although \dagf{}'s pruned loss is still the lowest. MUDDFormer, the closest dynamic design, loses the most at every size and sparsity.

No model relies on re-planning its routing: freezing \dagf{}'s predictor, or the baselines' routers, during pruning barely changes the result (Table~\ref{tab:pruning_full}). The advantage lies in the sources each head can draw on (Appendix~\ref{app:prunewhy}); with units removed at random both families lose much more and the gap shrinks at 75M (Appendix~\ref{app:prune}).

\section{Discussion}
\label{sec:discussion}

\paragraph{Token dependence.} Most of \dagf{}'s gain lies in the learned graph, a set of cross-layer connections and their weights, rather than in re-planning it for every token: the predictor's output can be replaced by its per-position mean at almost no cost (Appendix~\ref{sec:static}), and the predictor alone recovers about 90\% of the gain (Section~\ref{sec:source}). Input dependence, which dynamic cross-layer architectures cite to motivate their routers, matters mostly while the graph is learned: static weights trained directly (DenseFormer, static hyper-connections, an input-free table) recover little of it (Appendix~\ref{app:posthoc}). At inference the trained models still rely on the refinement's per-token corrections, and sparsifying the graph post hoc is costly (Appendix~\ref{app:topology}); distilling it into a fixed topology that a dense kernel can execute is future work.

\paragraph{Shared predictor and per-layer routers.} A shared predictor beats per-layer predictors that read the same tokens (Section~\ref{sec:source}), and \dagf{} has a lower loss than hyper-connections, mHC and MUDDFormer, whose per-layer routers read the hidden state, at every size but one (MUDDFormer at 150M; one seed per model; Table~\ref{tab:main}). At this scale, loss separates the two designs only partly. A separate module also makes the graph an object of its own: it can be averaged, frozen and patched, which is how Appendix~\ref{sec:static} locates the gain, and the pruned models degrade more gracefully. Whether the predictor keeps its lead at larger scale, for instance by planning routing that no single layer can see, is untested.

\paragraph{Inspecting the learned graph.} The predictor's output is a natural place to audit or steer an inference pass: it fixes which earlier results each head and layer combine before the backbone runs, and averaging, freezing and patching it are the causal interventions that generative interpretability asks of intermediate checkpoints \citep{yang2026generative}. We read the graph only in aggregate (Appendix~\ref{app:topology}); in the trained models the per-token variation lives mostly in the refinement, and whether the graph is meaningful to a human for a given input remains open. A larger predictor-only \dagf{} whose graph varies with the input in nameable ways, such as task or domain, could make the topology such a checkpoint, and routing among specialised modules (begun by the modular variant, Appendix~\ref{app:modular}) could make module outputs readable too.

\section{Related work}
\label{sec:related}

\paragraph{Cross-layer connectivity.} After highway and residual skips \citep{srivastava2015highway,he2016resnet}, DenseNet \citep{huang2017densenet} connected every layer to every earlier one, and transparent attention \citep{bapna2018transparent}, ELMo \citep{peters2018elmo}, DenseFormer \citep{pagliardini2024denseformer} and LAuReL \citep{menghani2024laurel} learn static weights over layer outputs. Hyper-connections \citep{zhu2024hyperconnections} widen the residual stream into several streams whose connections can depend on the incoming hidden state, and mHC \citep{mhc2026} constrains their residual-mixing matrices to be doubly stochastic. MUDDFormer \citep{muddformer2025} generates per-position query, key, value and residual weights from each layer's hidden state; \dagf{} shares this decomposition (per head for $q,k,v$) but plans the weights of all layers with one predictor over the tokens, which each layer refines per head from its input.

\paragraph{Conditional computation.} Conditional computation runs part of a network per input \citep{bengio2013estimating,han2022dynamic}: mixture-of-experts layers send each token to a few experts \citep{jacobs1991adaptive,shazeer2017moe,lepikhin2021gshard,fedus2022switch,zhou2022expertchoice}, adaptive-depth models halt or exit early \citep{graves2016adaptive,dehghani2019universal,elbayad2020depth,schuster2022confident}, and mixture-of-depths lets tokens skip whole blocks \citep{raposo2024mod}. BlockDrop \citep{wu2018blockdrop}, the closest analogue to a planned topology, picks all blocks an image skips with one policy network instead of per-layer gates \citep{wang2018skipnet,veit2020convolutional}. These methods choose \emph{how much} computation an input receives; \dagf{} keeps it fixed and chooses \emph{which earlier results} each layer reads.

\section{Conclusion}

We introduced \dagf{}, a Transformer whose cross-layer connections are planned by one predictor over the input tokens, trained jointly with the backbone. From 75M to 1B it lowers loss at every size and at equal compute, beats four routing baselines at every size but 150M, carries its gain to evaluation tasks and instruction finetuning, and degrades the least under structured pruning in five of six settings. Most of the gain lies in the learned graph, which the separate predictor makes an object that can be inspected and intervened on.

\section*{Limitations}
\label{sec:limitations}

\emph{Scale and seeds.} Our largest model has 1B parameters trained on 5 to 10B tokens, below its compute-optimal budget. Every pretraining run but the 300M pair is a single seed. Repeating the 300M pair with a second pretraining seed changes a model's held-out NLL by up to 0.08 nats per token (the vanilla Transformer on GSM8K solutions; at most 0.03 on WikiText-2; Appendix~\ref{app:pairs}), so smaller differences are noise; the pruning cells are single runs too. \emph{Cost.} Routing costs 1.18 to 2.13$\times$ the vanilla Transformer's FLOPs per token in the order used for training, and the trained models rely on the refinement's per-token corrections, so they cannot run with a fixed topology without retraining. \emph{Comparisons.} MUDDFormer trained with our recipe beats \dagf{} at 150M and loses at the other sizes, each a single seed, so loss alone does not settle that comparison; the 1B baselines are trained at 5B tokens only, the parameter-matched vanilla Transformer controls exist at 75M and 150M, and only the 75M pair was swept over learning rates. \emph{Evaluation.} Models of this size sit near chance on harder benchmarks, and a single LLM judge scored the chat and AlpacaEval comparisons. \emph{Interpretability and scale.} We intervene on the learned graph only as a whole and read it only in aggregate; whether its input-dependent part can be named, and whether the gains persist beyond 1B parameters, are left to follow-up work.

\ifanon\else
\section*{Acknowledgments}
We thank the INVITE Institute at the University of Illinois Urbana-Champaign and the ACCESS program, through the National Center for Supercomputing Applications (NCSA), for their support with computing credits.
This material is based upon work supported by the National Science Foundation and the Institute of Education Sciences under Grant \#2229612. Any opinions, findings, and conclusions or recommendations expressed in this material are those of the author(s) and do not necessarily reflect the views of National Science Foundation or the U.S. Department of Education.
This work used the Delta system at the National Center for Supercomputing Applications [award OAC 2005572] through allocations CIS251094 and CIS261897 from the Advanced Cyberinfrastructure Coordination Ecosystem: Services \& Support (ACCESS) program \citep{boerner2023access}, which is supported by U.S. National Science Foundation grants \#2138259, \#2138286, \#2138307, \#2137603, and \#2138296.
\fi

\bibliography{references}

\begin{thebibliography}{62}
\providecommand{\natexlab}[1]{#1}

\bibitem[{Allal et~al.(2025)Allal, Lozhkov, Bakouch, Bl{\'a}zquez, Penedo,
  Tunstall, Marafioti, Lajar{\'\i}n, Kydl{\'\i}{\v{c}}ek, Srivastav, Lochner,
  Fahlgren, Nguyen, Burtenshaw, Fourrier, Zhao, Larcher, Morlon, Zakka, Raffel,
  von Werra, and Wolf}]{allal2025smollm2}
Loubna~Ben Allal, Anton Lozhkov, Elie Bakouch, Gabriel~Mart{\'\i}n
  Bl{\'a}zquez, Guilherme Penedo, Lewis Tunstall, Andr{\'e}s Marafioti,
  Agust{\'\i}n~Piqueres Lajar{\'\i}n, Hynek Kydl{\'\i}{\v{c}}ek, Vaibhav
  Srivastav, Joshua Lochner, Caleb Fahlgren, Xuan-Son Nguyen, Ben Burtenshaw,
  Cl{\'e}mentine Fourrier, Haojun Zhao, Hugo Larcher, Mathieu Morlon, Cyril
  Zakka, and 3 others. 2025.
\newblock \href {https://openreview.net/forum?id=3JiCl2A14H} {{SmolLM2}: When
  smol goes big --- data-centric training of a fully open small language
  model}.
\newblock In \emph{Second Conference on Language Modeling (COLM)}.
\newblock ArXiv:2502.02737.

\bibitem[{Amini et~al.(2019)Amini, Gabriel, Lin, Koncel-Kedziorski, Choi, and
  Hajishirzi}]{amini2019mathqa}
Aida Amini, Saadia Gabriel, Shanchuan Lin, Rik Koncel-Kedziorski, Yejin Choi,
  and Hannaneh Hajishirzi. 2019.
\newblock \href {https://doi.org/10.18653/v1/N19-1245} {{MathQA}: Towards
  interpretable math word problem solving with operation-based formalisms}.
\newblock In \emph{Proceedings of the 2019 Conference of the North American
  Chapter of the Association for Computational Linguistics (NAACL)}, pages
  2357--2367.
\newblock ArXiv:1905.13319.

\bibitem[{Bapna et~al.(2018)Bapna, Chen, Firat, Cao, and
  Wu}]{bapna2018transparent}
Ankur Bapna, Mia~Xu Chen, Orhan Firat, Yuan Cao, and Yonghui Wu. 2018.
\newblock Training deeper neural machine translation models with transparent
  attention.
\newblock In \emph{Proceedings of the 2018 Conference on Empirical Methods in
  Natural Language Processing (EMNLP)}.
\newblock ArXiv:1808.07561.

\bibitem[{Bengio et~al.(2013)Bengio, L{\'e}onard, and
  Courville}]{bengio2013estimating}
Yoshua Bengio, Nicholas L{\'e}onard, and Aaron~C. Courville. 2013.
\newblock Estimating or propagating gradients through stochastic neurons for
  conditional computation.
\newblock \emph{arXiv preprint arXiv:1308.3432}.

\bibitem[{Bisk et~al.(2020)Bisk, Zellers, Bras, Gao, and Choi}]{bisk2020piqa}
Yonatan Bisk, Rowan Zellers, Ronan~Le Bras, Jianfeng Gao, and Yejin Choi. 2020.
\newblock {PIQA}: Reasoning about physical commonsense in natural language.
\newblock In \emph{Proceedings of the AAAI Conference on Artificial
  Intelligence}.
\newblock ArXiv:1911.11641.

\bibitem[{Boerner et~al.(2023)Boerner, Deems, Furlani, Knuth, and
  Towns}]{boerner2023access}
Timothy~J. Boerner, Stephen Deems, Thomas~R. Furlani, Shelley~L. Knuth, and
  John Towns. 2023.
\newblock \href {https://doi.org/10.1145/3569951.3597559} {{ACCESS}: Advancing
  innovation: {NSF}'s advanced cyberinfrastructure coordination ecosystem:
  Services \& support}.
\newblock In \emph{Practice and Experience in Advanced Research Computing
  ({PEARC} '23)}, pages 173--176. ACM.

\bibitem[{Clark et~al.(2019)Clark, Lee, Chang, Kwiatkowski, Collins, and
  Toutanova}]{clark2019boolq}
Christopher Clark, Kenton Lee, Ming-Wei Chang, Tom Kwiatkowski, Michael
  Collins, and Kristina Toutanova. 2019.
\newblock {BoolQ}: Exploring the surprising difficulty of natural yes/no
  questions.
\newblock In \emph{Proceedings of the 2019 Conference of the North American
  Chapter of the Association for Computational Linguistics (NAACL)}.
\newblock ArXiv:1905.10044.

\bibitem[{Clark et~al.(2018)Clark, Cowhey, Etzioni, Khot, Sabharwal, Schoenick,
  and Tafjord}]{clark2018arc}
Peter Clark, Isaac Cowhey, Oren Etzioni, Tushar Khot, Ashish Sabharwal, Carissa
  Schoenick, and Oyvind Tafjord. 2018.
\newblock Think you have solved question answering? {Try} {ARC}, the {AI2}
  reasoning challenge.
\newblock ArXiv:1803.05457.

\bibitem[{Cobbe et~al.(2021)Cobbe, Kosaraju, Bavarian, Chen, Jun, Kaiser,
  Plappert, Tworek, Hilton, Nakano, Hesse, and Schulman}]{cobbe2021gsm8k}
Karl Cobbe, Vineet Kosaraju, Mohammad Bavarian, Mark Chen, Heewoo Jun, Lukasz
  Kaiser, Matthias Plappert, Jerry Tworek, Jacob Hilton, Reiichiro Nakano,
  Christopher Hesse, and John Schulman. 2021.
\newblock Training verifiers to solve math word problems.
\newblock ArXiv:2110.14168.

\bibitem[{Conover et~al.(2023)Conover, Hayes, Mathur, Xie, Wan, Shah, Ghodsi,
  Wendell, Zaharia, and Xin}]{conover2023dolly}
Mike Conover, Matt Hayes, Ankit Mathur, Jianwei Xie, Jun Wan, Sam Shah, Ali
  Ghodsi, Patrick Wendell, Matei Zaharia, and Reynold Xin. 2023.
\newblock \href
  {https://www.databricks.com/blog/2023/04/12/dolly-first-open-commercially-viable-instruction-tuned-llm}
  {Free dolly: Introducing the world's first truly open instruction-tuned
  {LLM}}.
\newblock Databricks blog, April 12, 2023.

\bibitem[{Dehghani et~al.(2019)Dehghani, Gouws, Vinyals, Uszkoreit, and
  Kaiser}]{dehghani2019universal}
Mostafa Dehghani, Stephan Gouws, Oriol Vinyals, Jakob Uszkoreit, and {\L}ukasz
  Kaiser. 2019.
\newblock Universal transformers.
\newblock In \emph{International Conference on Learning Representations}.

\bibitem[{Elbayad et~al.(2020)Elbayad, Gu, Grave, and Auli}]{elbayad2020depth}
Maha Elbayad, Jiatao Gu, Edouard Grave, and Michael Auli. 2020.
\newblock Depth-adaptive transformer.
\newblock In \emph{International Conference on Learning Representations}.

\bibitem[{Elhage et~al.(2021)Elhage, Nanda, Olsson, Henighan, Joseph, Mann,
  Askell, Bai, Chen, Conerly, DasSarma, Drain, Ganguli, Hatfield-Dodds,
  Hernandez, Jones, Kernion, Lovitt, Ndousse, Amodei, Brown, Clark, Kaplan,
  McCandlish, and Olah}]{elhage2021}
Nelson Elhage, Neel Nanda, Catherine Olsson, Tom Henighan, Nicholas Joseph, Ben
  Mann, Amanda Askell, Yuntao Bai, Anna Chen, Tom Conerly, Nova DasSarma, Dawn
  Drain, Deep Ganguli, Zac Hatfield-Dodds, Danny Hernandez, Andy Jones, Jackson
  Kernion, Liane Lovitt, Kamal Ndousse, and 6 others. 2021.
\newblock \href {https://transformer-circuits.pub/2021/framework/index.html} {A
  mathematical framework for transformer circuits}.
\newblock Transformer Circuits Thread.

\bibitem[{Fedus et~al.(2022)Fedus, Zoph, and Shazeer}]{fedus2022switch}
William Fedus, Barret Zoph, and Noam Shazeer. 2022.
\newblock \href {http://jmlr.org/papers/v23/21-0998.html} {Switch transformers:
  Scaling to trillion parameter models with simple and efficient sparsity}.
\newblock \emph{Journal of Machine Learning Research}, 23(120):1--39.
\newblock ArXiv:2101.03961.

\bibitem[{Gao et~al.(2024)Gao, Tow, Abbasi, Biderman, Black, DiPofi, Foster,
  Golding, Hsu, Noac'h, Li, McDonell, Muennighoff, Ociepa, Phang, Reynolds,
  Schoelkopf, Skowron, Sutawika, Tang, Thite, Wang, Wang, and
  Zou}]{gao2024lmeval}
Leo Gao, Jonathan Tow, Baber Abbasi, Stella Biderman, Sid Black, Anthony
  DiPofi, Charles Foster, Laurence Golding, Jeffrey Hsu, Alain~Le Noac'h,
  Haonan Li, Kyle McDonell, Niklas Muennighoff, Chris Ociepa, Jason Phang,
  Laria Reynolds, Hailey Schoelkopf, Aviya Skowron, Lintang Sutawika, and 5
  others. 2024.
\newblock \href {https://doi.org/10.5281/zenodo.12608602} {The language model
  evaluation harness}.
\newblock Version v0.4.3.

\bibitem[{Graves(2016)}]{graves2016adaptive}
Alex Graves. 2016.
\newblock Adaptive computation time for recurrent neural networks.
\newblock \emph{arXiv preprint arXiv:1603.08983}.

\bibitem[{Han et~al.(2022)Han, Huang, Song, Yang, Wang, and
  Wang}]{han2022dynamic}
Yizeng Han, Gao Huang, Shiji Song, Le~Yang, Honghui Wang, and Yulin Wang. 2022.
\newblock \href {https://doi.org/10.1109/TPAMI.2021.3117837} {Dynamic neural
  networks: A survey}.
\newblock \emph{IEEE Transactions on Pattern Analysis and Machine
  Intelligence}, 44(11):7436--7456.

\bibitem[{He et~al.(2016)He, Zhang, Ren, and Sun}]{he2016resnet}
Kaiming He, Xiangyu Zhang, Shaoqing Ren, and Jian Sun. 2016.
\newblock Deep residual learning for image recognition.
\newblock In \emph{IEEE Conference on Computer Vision and Pattern Recognition
  (CVPR)}.
\newblock ArXiv:1512.03385.

\bibitem[{Hoffmann et~al.(2022)Hoffmann, Borgeaud, Mensch, Buchatskaya, Cai,
  Rutherford, de~Las~Casas, Hendricks, Welbl, Clark, Hennigan, Noland,
  Millican, van~den Driessche, Damoc, Guy, Osindero, Simonyan, Elsen, Vinyals,
  Rae, and Sifre}]{hoffmann2022chinchilla}
Jordan Hoffmann, Sebastian Borgeaud, Arthur Mensch, Elena Buchatskaya, Trevor
  Cai, Eliza Rutherford, Diego de~Las~Casas, Lisa~Anne Hendricks, Johannes
  Welbl, Aidan Clark, Tom Hennigan, Eric Noland, Katie Millican, George van~den
  Driessche, Bogdan Damoc, Aurelia Guy, Simon Osindero, Karen Simonyan, Erich
  Elsen, and 3 others. 2022.
\newblock \href {https://doi.org/10.52202/068431-2176} {Training
  compute-optimal large language models}.
\newblock In \emph{Advances in Neural Information Processing Systems},
  volume~35, pages 30016--30030.
\newblock ArXiv:2203.15556.

\bibitem[{Huang et~al.(2017)Huang, Liu, van~der Maaten, and
  Weinberger}]{huang2017densenet}
Gao Huang, Zhuang Liu, Laurens van~der Maaten, and Kilian~Q. Weinberger. 2017.
\newblock Densely connected convolutional networks.
\newblock In \emph{IEEE Conference on Computer Vision and Pattern Recognition
  (CVPR)}.
\newblock ArXiv:1608.06993.

\bibitem[{Jacobs et~al.(1991)Jacobs, Jordan, Nowlan, and
  Hinton}]{jacobs1991adaptive}
Robert~A. Jacobs, Michael~I. Jordan, Steven~J. Nowlan, and Geoffrey~E. Hinton.
  1991.
\newblock \href {https://doi.org/10.1162/neco.1991.3.1.79} {Adaptive mixtures
  of local experts}.
\newblock \emph{Neural Computation}, 3(1):79--87.

\bibitem[{Kaplan et~al.(2020)Kaplan, McCandlish, Henighan, Brown, Chess, Child,
  Gray, Radford, Wu, and Amodei}]{kaplan2020scaling}
Jared Kaplan, Sam McCandlish, Tom Henighan, Tom~B. Brown, Benjamin Chess, Rewon
  Child, Scott Gray, Alec Radford, Jeffrey Wu, and Dario Amodei. 2020.
\newblock Scaling laws for neural language models.
\newblock ArXiv:2001.08361.

\bibitem[{Lepikhin et~al.(2021)Lepikhin, Lee, Xu, Chen, Firat, Huang, Krikun,
  Shazeer, and Chen}]{lepikhin2021gshard}
Dmitry Lepikhin, HyoukJoong Lee, Yuanzhong Xu, Dehao Chen, Orhan Firat, Yanping
  Huang, Maxim Krikun, Noam Shazeer, and Zhifeng Chen. 2021.
\newblock {GShard}: Scaling giant models with conditional computation and
  automatic sharding.
\newblock In \emph{International Conference on Learning Representations}.

\bibitem[{Li et~al.(2023)Li, Zhang, Dubois, Taori, Gulrajani, Guestrin, Liang,
  and Hashimoto}]{alpaca_eval}
Xuechen Li, Tianyi Zhang, Yann Dubois, Rohan Taori, Ishaan Gulrajani, Carlos
  Guestrin, Percy Liang, and Tatsunori~B. Hashimoto. 2023.
\newblock \href {https://github.com/tatsu-lab/alpaca_eval} {Alpacaeval: An
  automatic evaluator of instruction-following models}.
\newblock GitHub repository.

\bibitem[{Meng et~al.(2022)Meng, Bau, Andonian, and Belinkov}]{meng2022rome}
Kevin Meng, David Bau, Alex Andonian, and Yonatan Belinkov. 2022.
\newblock Locating and editing factual associations in {GPT}.
\newblock In \emph{Advances in Neural Information Processing Systems}.
\newblock ArXiv:2202.05262.

\bibitem[{Menghani et~al.(2025)Menghani, Kumar, and Kumar}]{menghani2024laurel}
Gaurav Menghani, Ravi Kumar, and Sanjiv Kumar. 2025.
\newblock {LAuReL}: Learned augmented residual layer.
\newblock In \emph{Proceedings of the 42nd International Conference on Machine
  Learning}, volume 267 of \emph{Proceedings of Machine Learning Research},
  pages 43826--43836. PMLR.
\newblock ArXiv:2411.07501.

\bibitem[{Merity et~al.(2017)Merity, Xiong, Bradbury, and
  Socher}]{merity2016wikitext}
Stephen Merity, Caiming Xiong, James Bradbury, and Richard Socher. 2017.
\newblock \href {https://openreview.net/forum?id=Byj72udxe} {Pointer sentinel
  mixture models}.
\newblock In \emph{International Conference on Learning Representations
  (ICLR)}.
\newblock ArXiv:1609.07843.

\bibitem[{Michel et~al.(2019)Michel, Levy, and Neubig}]{michel2019heads}
Paul Michel, Omer Levy, and Graham Neubig. 2019.
\newblock Are sixteen heads really better than one?
\newblock In \emph{Advances in Neural Information Processing Systems}.
\newblock ArXiv:1905.10650.

\bibitem[{Mihaylov et~al.(2018)Mihaylov, Clark, Khot, and
  Sabharwal}]{mihaylov2018openbookqa}
Todor Mihaylov, Peter Clark, Tushar Khot, and Ashish Sabharwal. 2018.
\newblock Can a suit of armor conduct electricity? {A} new dataset for open
  book question answering.
\newblock In \emph{Proceedings of the 2018 Conference on Empirical Methods in
  Natural Language Processing (EMNLP)}.
\newblock ArXiv:1809.02789.

\bibitem[{Molchanov et~al.(2019)Molchanov, Mallya, Tyree, Frosio, and
  Kautz}]{molchanov2019taylor}
Pavlo Molchanov, Arun Mallya, Stephen Tyree, Iuri Frosio, and Jan Kautz. 2019.
\newblock \href {https://doi.org/10.1109/CVPR.2019.01152} {Importance
  estimation for neural network pruning}.
\newblock In \emph{IEEE/CVF Conference on Computer Vision and Pattern
  Recognition (CVPR)}.
\newblock ArXiv:1906.10771.

\bibitem[{Pagliardini et~al.(2024)Pagliardini, Mohtashami, Fleuret, and
  Jaggi}]{pagliardini2024denseformer}
Matteo Pagliardini, Amirkeivan Mohtashami, Fran{\c{c}}ois Fleuret, and Martin
  Jaggi. 2024.
\newblock \href {https://doi.org/10.52202/079017-4336} {{DenseFormer}:
  Enhancing information flow in transformers via depth weighted averaging}.
\newblock In \emph{Advances in Neural Information Processing Systems},
  volume~37, pages 136479--136508. Curran Associates, Inc.
\newblock ArXiv:2402.02622.

\bibitem[{Paperno et~al.(2016)Paperno, Kruszewski, Lazaridou, Pham, Bernardi,
  Pezzelle, Baroni, Boleda, and Fern{\'a}ndez}]{paperno2016lambada}
Denis Paperno, Germ{\'a}n Kruszewski, Angeliki Lazaridou, Quan~Ngoc Pham,
  Raffaella Bernardi, Sandro Pezzelle, Marco Baroni, Gemma Boleda, and Raquel
  Fern{\'a}ndez. 2016.
\newblock The {LAMBADA} dataset: Word prediction requiring a broad discourse
  context.
\newblock In \emph{Proceedings of the 54th Annual Meeting of the Association
  for Computational Linguistics (ACL)}.
\newblock ArXiv:1606.06031.

\bibitem[{Peters et~al.(2018)Peters, Neumann, Iyyer, Gardner, Clark, Lee, and
  Zettlemoyer}]{peters2018elmo}
Matthew~E. Peters, Mark Neumann, Mohit Iyyer, Matt Gardner, Christopher Clark,
  Kenton Lee, and Luke Zettlemoyer. 2018.
\newblock Deep contextualized word representations.
\newblock In \emph{Proceedings of the 2018 Conference of the North American
  Chapter of the Association for Computational Linguistics (NAACL)}.
\newblock ArXiv:1802.05365.

\bibitem[{{Qwen Team}(2025)}]{qwen3coder}
{Qwen Team}. 2025.
\newblock \href {https://huggingface.co/Qwen/Qwen3-Coder-30B-A3B-Instruct}
  {{Qwen3-Coder-30B-A3B-Instruct} model card}.
\newblock Hugging Face.

\bibitem[{Raposo et~al.(2024)Raposo, Ritter, Richards, Lillicrap, Humphreys,
  and Santoro}]{raposo2024mod}
David Raposo, Sam Ritter, Blake Richards, Timothy Lillicrap, Peter~Conway
  Humphreys, and Adam Santoro. 2024.
\newblock Mixture-of-depths: Dynamically allocating compute in
  transformer-based language models.
\newblock ArXiv:2404.02258.

\bibitem[{Sakaguchi et~al.(2020)Sakaguchi, Bras, Bhagavatula, and
  Choi}]{sakaguchi2020winogrande}
Keisuke Sakaguchi, Ronan~Le Bras, Chandra Bhagavatula, and Yejin Choi. 2020.
\newblock \href {https://doi.org/10.1609/aaai.v34i05.6399} {{WinoGrande}: An
  adversarial {Winograd} schema challenge at scale}.
\newblock In \emph{Proceedings of the AAAI Conference on Artificial
  Intelligence}, volume~34, pages 8732--8740.
\newblock ArXiv:1907.10641.

\bibitem[{Sap et~al.(2019)Sap, Rashkin, Chen, Bras, and
  Choi}]{sap2019socialiqa}
Maarten Sap, Hannah Rashkin, Derek Chen, Ronan~Le Bras, and Yejin Choi. 2019.
\newblock \href {https://doi.org/10.18653/v1/D19-1454} {{Social IQa}:
  Commonsense reasoning about social interactions}.
\newblock In \emph{Proceedings of the 2019 Conference on Empirical Methods in
  Natural Language Processing and the 9th International Joint Conference on
  Natural Language Processing (EMNLP-IJCNLP)}, pages 4463--4473.
\newblock ArXiv:1904.09728.

\bibitem[{Schuster et~al.(2022)Schuster, Fisch, Gupta, Dehghani, Bahri, Tran,
  Tay, and Metzler}]{schuster2022confident}
Tal Schuster, Adam Fisch, Jai Gupta, Mostafa Dehghani, Dara Bahri, Vinh~Q.
  Tran, Yi~Tay, and Donald Metzler. 2022.
\newblock Confident adaptive language modeling.
\newblock In \emph{Advances in Neural Information Processing Systems},
  volume~35.

\bibitem[{Shazeer et~al.(2017)Shazeer, Mirhoseini, Maziarz, Davis, Le, Hinton,
  and Dean}]{shazeer2017moe}
Noam Shazeer, Azalia Mirhoseini, Krzysztof Maziarz, Andy Davis, Quoc Le,
  Geoffrey Hinton, and Jeff Dean. 2017.
\newblock Outrageously large neural networks: The sparsely-gated
  mixture-of-experts layer.
\newblock In \emph{International Conference on Learning Representations
  (ICLR)}.
\newblock ArXiv:1701.06538.

\bibitem[{Soldaini et~al.(2024)Soldaini, Kinney, Bhagia, Schwenk, Atkinson,
  Authur, Bogin, Chandu, Dumas, Elazar, Hofmann, Jha, Kumar, Lucy, Lyu,
  Lambert, Magnusson, Morrison, Muennighoff, Naik, Nam, Peters, Ravichander,
  Richardson, Shen, Strubell, Subramani, Tafjord, Walsh, Zettlemoyer, Smith,
  Hajishirzi, Beltagy, Groeneveld, Dodge, and Lo}]{soldaini2024dolma}
Luca Soldaini, Rodney Kinney, Akshita Bhagia, Dustin Schwenk, David Atkinson,
  Russell Authur, Ben Bogin, Khyathi Chandu, Jennifer Dumas, Yanai Elazar,
  Valentin Hofmann, Ananya~Harsh Jha, Sachin Kumar, Li~Lucy, Xinxi Lyu, Nathan
  Lambert, Ian Magnusson, Jacob Morrison, Niklas Muennighoff, and 17 others.
  2024.
\newblock \href {https://doi.org/10.18653/v1/2024.acl-long.840} {Dolma: an open
  corpus of three trillion tokens for language model pretraining research}.
\newblock In \emph{Proceedings of the 62nd Annual Meeting of the Association
  for Computational Linguistics (Volume 1: Long Papers)}, pages 15725--15788,
  Bangkok, Thailand. Association for Computational Linguistics.
\newblock ArXiv:2402.00159.

\bibitem[{Srivastava et~al.(2015)Srivastava, Greff, and
  Schmidhuber}]{srivastava2015highway}
Rupesh~Kumar Srivastava, Klaus Greff, and J{\"u}rgen Schmidhuber. 2015.
\newblock Training very deep networks.
\newblock In \emph{Advances in Neural Information Processing Systems},
  volume~28.
\newblock ArXiv:1507.06228; extends arXiv:1505.00387 (Highway Networks).

\bibitem[{Talmor et~al.(2019)Talmor, Herzig, Lourie, and
  Berant}]{talmor2019commonsenseqa}
Alon Talmor, Jonathan Herzig, Nicholas Lourie, and Jonathan Berant. 2019.
\newblock {CommonsenseQA}: A question answering challenge targeting commonsense
  knowledge.
\newblock In \emph{Proceedings of the 2019 Conference of the North American
  Chapter of the Association for Computational Linguistics (NAACL)}.
\newblock ArXiv:1811.00937.

\bibitem[{Taori et~al.(2023)Taori, Gulrajani, Zhang, Dubois, Li, Guestrin,
  Liang, and Hashimoto}]{taori2023alpaca}
Rohan Taori, Ishaan Gulrajani, Tianyi Zhang, Yann Dubois, Xuechen Li, Carlos
  Guestrin, Percy Liang, and Tatsunori~B. Hashimoto. 2023.
\newblock \href {https://github.com/tatsu-lab/stanford_alpaca} {Stanford
  alpaca: An instruction-following {LLaMA} model}.
\newblock GitHub repository.

\bibitem[{Vaswani et~al.(2017)Vaswani, Shazeer, Parmar, Uszkoreit, Jones,
  Gomez, Kaiser, and Polosukhin}]{vaswani2017}
Ashish Vaswani, Noam Shazeer, Niki Parmar, Jakob Uszkoreit, Llion Jones,
  Aidan~N. Gomez, {\L}ukasz Kaiser, and Illia Polosukhin. 2017.
\newblock Attention is all you need.
\newblock In \emph{Advances in Neural Information Processing Systems}.
\newblock ArXiv:1706.03762.

\bibitem[{Veit and Belongie(2020)}]{veit2020convolutional}
Andreas Veit and Serge Belongie. 2020.
\newblock \href {https://doi.org/10.1007/s11263-019-01190-4} {Convolutional
  networks with adaptive inference graphs}.
\newblock \emph{International Journal of Computer Vision}, 128(3):730--741.

\bibitem[{Vig et~al.(2020)Vig, Gehrmann, Belinkov, Qian, Nevo, Singer, and
  Shieber}]{vig2020causal}
Jesse Vig, Sebastian Gehrmann, Yonatan Belinkov, Sharon Qian, Daniel Nevo,
  Yaron Singer, and Stuart Shieber. 2020.
\newblock Investigating gender bias in language models using causal mediation
  analysis.
\newblock In \emph{Advances in Neural Information Processing Systems}.

\bibitem[{Walsh et~al.(2025)Walsh, Soldaini, Groeneveld, Lo, Arora, Bhagia, Gu,
  Huang, Jordan, Lambert, Schwenk, Tafjord, Anderson, Atkinson, Brahman, Clark,
  Dasigi, Dziri, Ettinger, Guerquin, Heineman, Ivison, Koh, Liu, Malik,
  Merrill, Miranda, Morrison, Murray, Nam, Poznanski, Pyatkin, Rangapur,
  Schmitz, Skjonsberg, Wadden, Wilhelm, Wilson, Zettlemoyer, Farhadi, Smith,
  and Hajishirzi}]{olmo2}
Evan~Pete Walsh, Luca Soldaini, Dirk Groeneveld, Kyle Lo, Shane Arora, Akshita
  Bhagia, Yuling Gu, Shengyi Huang, Matt Jordan, Nathan Lambert, Dustin
  Schwenk, Oyvind Tafjord, Taira Anderson, David Atkinson, Faeze Brahman,
  Christopher Clark, Pradeep Dasigi, Nouha Dziri, Allyson Ettinger, and 23
  others. 2025.
\newblock \href {https://openreview.net/forum?id=2ezugTT9kU} {2 {OLMo} 2
  furious ({COLM}'s version)}.
\newblock In \emph{Conference on Language Modeling (COLM)}.
\newblock Extended version: arXiv:2501.00656.

\bibitem[{Wang et~al.(2018)Wang, Yu, Dou, Darrell, and
  Gonzalez}]{wang2018skipnet}
Xin Wang, Fisher Yu, Zi-Yi Dou, Trevor Darrell, and Joseph~E. Gonzalez. 2018.
\newblock \href {https://doi.org/10.1007/978-3-030-01261-8_25} {{SkipNet}:
  Learning dynamic routing in convolutional networks}.
\newblock In \emph{Computer Vision -- ECCV 2018}, pages 420--436.

\bibitem[{Welbl et~al.(2017)Welbl, Liu, and Gardner}]{welbl2017sciq}
Johannes Welbl, Nelson~F. Liu, and Matt Gardner. 2017.
\newblock Crowdsourcing multiple choice science questions.
\newblock In \emph{Proceedings of the 3rd Workshop on Noisy User-generated Text
  (W-NUT)}.
\newblock ArXiv:1707.06209.

\bibitem[{Wu et~al.(2018)Wu, Nagarajan, Kumar, Rennie, Davis, Grauman, and
  Feris}]{wu2018blockdrop}
Zuxuan Wu, Tushar Nagarajan, Abhishek Kumar, Steven Rennie, Larry~S. Davis,
  Kristen Grauman, and Rogerio Feris. 2018.
\newblock \href {https://doi.org/10.1109/CVPR.2018.00919} {{BlockDrop}: Dynamic
  inference paths in residual networks}.
\newblock In \emph{Proceedings of the IEEE/CVF Conference on Computer Vision
  and Pattern Recognition}, pages 8817--8826.

\bibitem[{Xia et~al.(2022)Xia, Zhong, and Chen}]{xia2022cofi}
Mengzhou Xia, Zexuan Zhong, and Danqi Chen. 2022.
\newblock Structured pruning learns compact and accurate models.
\newblock In \emph{Proceedings of the 60th Annual Meeting of the Association
  for Computational Linguistics (ACL)}.
\newblock ArXiv:2204.00408.

\bibitem[{Xiao et~al.(2025)Xiao, Meng, Li, and Yuan}]{muddformer2025}
Da~Xiao, Qingye Meng, Shengping Li, and Xingyuan Yuan. 2025.
\newblock {MUDDFormer}: Breaking residual bottlenecks in transformers via
  multiway dynamic dense connections.
\newblock In \emph{Proceedings of the 42nd International Conference on Machine
  Learning (ICML)}, volume 267 of \emph{Proceedings of Machine Learning
  Research}, pages 68440--68458. PMLR.
\newblock ArXiv:2502.12170.

\bibitem[{Xie et~al.(2026)Xie, Wei, Cao, Zhao, Deng, Li, Dai, Gao, Chang, Yu,
  Zhao, Zhou, Xu, Zhang, Zeng, Hu, Wang, Yuan, Wang, and Liang}]{mhc2026}
Zhenda Xie, Yixuan Wei, Huanqi Cao, Chenggang Zhao, Chengqi Deng, Jiashi Li,
  Damai Dai, Huazuo Gao, Jiang Chang, Kuai Yu, Liang Zhao, Shangyan Zhou, Zhean
  Xu, Zhengyan Zhang, Wangding Zeng, Shengding Hu, Yuqing Wang, Jingyang Yuan,
  Lean Wang, and Wenfeng Liang. 2026.
\newblock {mHC}: Manifold-constrained hyper-connections.
\newblock ArXiv:2512.24880.

\bibitem[{Yang et~al.(2025)Yang, Li, Yang, Zhang, Hui, Zheng et~al.}]{qwen3}
An~Yang, Anfeng Li, Baosong Yang, Beichen Zhang, Binyuan Hui, Bo~Zheng, et~al.
  2025.
\newblock {Qwen3} technical report.
\newblock ArXiv:2505.09388.

\bibitem[{Yang(2026)}]{yang2026generative}
Xiaocong Yang. 2026.
\newblock \href {https://arxiv.org/abs/2609.13529} {Generative interpretability
  via scalable neuro-symbolic models}.
\newblock In \emph{ACM AI Summit 2026 (AI Summit '26)}, Atlanta, GA, USA. ACM.

\bibitem[{Yue et~al.(2024)Yue, Qu, Zhang, Fu, Huang, Sun, Su, and
  Chen}]{yue2023mammoth}
Xiang Yue, Xingwei Qu, Ge~Zhang, Yao Fu, Wenhao Huang, Huan Sun, Yu~Su, and
  Wenhu Chen. 2024.
\newblock \href
  {https://proceedings.iclr.cc/paper_files/paper/2024/hash/b063829b922fdeb4fa3472dd3471ff43-Abstract-Conference.html}
  {{MAmmoTH}: Building math generalist models through hybrid instruction
  tuning}.
\newblock In \emph{International Conference on Learning Representations
  (ICLR)}, pages 40320--40341.
\newblock ArXiv:2309.05653.

\bibitem[{Zellers et~al.(2019)Zellers, Holtzman, Bisk, Farhadi, and
  Choi}]{zellers2019hellaswag}
Rowan Zellers, Ari Holtzman, Yonatan Bisk, Ali Farhadi, and Yejin Choi. 2019.
\newblock {HellaSwag}: Can a machine really finish your sentence?
\newblock In \emph{Proceedings of the 57th Annual Meeting of the Association
  for Computational Linguistics (ACL)}.
\newblock ArXiv:1905.07830.

\bibitem[{Zhang and Nanda(2024)}]{zhang2024patching}
Fred Zhang and Neel Nanda. 2024.
\newblock Towards best practices of activation patching in language models:
  Metrics and methods.
\newblock In \emph{International Conference on Learning Representations
  (ICLR)}.
\newblock ArXiv:2309.16042.

\bibitem[{Zheng et~al.(2023)Zheng, Chiang, Sheng, Zhuang, Wu, Zhuang, Lin, Li,
  Li, Xing, Zhang, Gonzalez, and Stoica}]{zheng2023judge}
Lianmin Zheng, Wei-Lin Chiang, Ying Sheng, Siyuan Zhuang, Zhanghao Wu, Yonghao
  Zhuang, Zi~Lin, Zhuohan Li, Dacheng Li, Eric~P. Xing, Hao Zhang, Joseph~E.
  Gonzalez, and Ion Stoica. 2023.
\newblock Judging {LLM}-as-a-judge with {MT}-bench and chatbot arena.
\newblock In \emph{Advances in Neural Information Processing Systems (Datasets
  and Benchmarks Track)}.
\newblock ArXiv:2306.05685.

\bibitem[{Zhou et~al.(2022)Zhou, Lei, Liu, Du, Huang, Zhao, Dai, Chen, Le, and
  Laudon}]{zhou2022expertchoice}
Yanqi Zhou, Tao Lei, Hanxiao Liu, Nan Du, Yanping Huang, Vincent Zhao,
  Andrew~M. Dai, Zhifeng Chen, Quoc~V. Le, and James Laudon. 2022.
\newblock \href {https://doi.org/10.52202/068431-0515} {Mixture-of-experts with
  expert choice routing}.
\newblock In \emph{Advances in Neural Information Processing Systems},
  volume~35.

\bibitem[{Zhu et~al.(2025)Zhu, Huang, Huang, Zeng, Mao, Wu, Min, and
  Zhou}]{zhu2024hyperconnections}
Defa Zhu, Hongzhi Huang, Zihao Huang, Yutao Zeng, Yunyao Mao, Banggu Wu, Qiyang
  Min, and Xun Zhou. 2025.
\newblock Hyper-connections.
\newblock In \emph{International Conference on Learning Representations
  (ICLR)}, pages 97183--97219.
\newblock ArXiv:2409.19606.

\bibitem[{Zhu and Gupta(2018)}]{zhu2017prune}
Michael Zhu and Suyog Gupta. 2018.
\newblock To prune, or not to prune: Exploring the efficacy of pruning for
  model compression.
\newblock In \emph{International Conference on Learning Representations (ICLR),
  Workshop Track}.
\newblock ArXiv:1710.01878.

\end{thebibliography}

\appendix

\section{Compute accounting}
\label{app:flops}
Forward MACs per token for the vanilla Transformer backbone are $\sum_{l=1}^{L}(4D^2 + 3DI) + VD$ plus attention over the context ($2TD$ per layer at sequence length $T$), with $D$ the model width, $I$ the MLP width and $V$ the vocabulary. In the project-then-mix order used in training, \dagf{} adds, at every routed layer $l\ge2$, the projection of $l{-}1$ extra sources through the fused $q,k,v$ weights ($(l{-}1)\cdot 3D^2$), the per-head mixing sums ($l(3D + D)$), the predictor (embedding lookup, 2 encoder layers of width 256, trunk and per-layer heads) and the local refinement networks ($D\cdot 128 + 128\cdot(3H{+}1)\,l$ per layer). Training FLOPs are $6\times$ the forward MACs times tokens: the $C \approx 6N$ per token estimate of \citet{kaplan2020scaling} (a backward pass costs about twice the forward), applied to all the forward MACs above, so the unembedding, attention-context and routing terms that $6N$ leaves out are included. The ratios are 1.18 (75M), 1.33 (150M), 1.65 (300M) and 2.13 (1B); the predictor itself is under 2\% of the total, the source projections dominate. Table~\ref{tab:traincost} gives the measured training cost of every model of Table~\ref{tab:main}: forward and backward passes over 2 sequences of 1024 tokens with bf16 weights, as in pretraining, on one A100-40GB, every model compiled with torch.compile (median of 20 iterations after warm-up, no optimizer step). Every cell, here and in Table~\ref{tab:main}, is the fastest of two runs; the runs agree within 1\% in every cell but one, the 75M vanilla Transformer, timed first in its job (6\%). The baselines are measured at random initialization, since the weights do not change the speed. \dagf{} trains at 1.44 to 1.77$\times$ the vanilla Transformer's time in the faster of its two orders; under 300M mHC is the slowest to train. Replacing the predictor by a constant table changes \dagf{}'s time by 1 to 3\%. At inference \dagf{} keeps every layer's output for the later layers to mix, so its peak memory for 8 sequences of 1024 tokens is 1.9 vs 1.8 GB (75M) to 9.1 vs 4.2 GB (1B), close to MUDDFormer, which also keeps them (8.7 GB at 1B). Without compilation every routed model is slower, the baselines most (hyper-connections 2.3 to 2.8$\times$, mHC 2.3 to 3.2$\times$ at inference; \texttt{experiments/scaling/inference\_cost.json}).

\subsection{Two exact computation orders}
\label{app:orders}
The $q,k,v$ projections have no bias, so for each head $h$ the projection of a mixture equals the mixture of projections, $W_q^h \sum_{j<l} \alpha^{(l)}_{q,h,j} X_j = \sum_{j<l} \alpha^{(l)}_{q,h,j} W_q^h X_j$ (likewise for $k$ and $v$), and \dagf{} can compute a layer's inputs in either order. Project-then-mix, used in training, runs one fused $q,k,v$ GEMM over the $l$ stacked sources: $(l{-}1)\cdot 3D^2$ extra MACs at layer $l$, at full GEMM efficiency. Mix-then-project first forms one mixture of the $l$ sources per head for each of $q,k,v$ ($3HlD$ MACs) and then applies each head's slice of the projection, so the whole model needs only 1.02 to 1.04$\times$ the vanilla Transformer's FLOPs; in exchange it writes and reads $3H$ vectors of width $D$ per token and layer and runs $H$ narrow GEMMs instead of one wide one, so it is bound by memory traffic. Per token and layer, the first order spends about $D/H$ extra MACs, the head width, for every value the second order moves: 64 up to 300M and 128 at 1B. Consistent with this, projecting first is faster at inference up to 300M (1.49 vs 1.68$\times$ the vanilla Transformer at 75M, 1.61 vs 1.77 at 150M, 1.89 vs 1.93 at 300M) and mixing first at 1B (1.81 vs 2.38$\times$); in training the two are within 3\% up to 300M and mixing first is faster at 1B (1.65 vs 2.17$\times$). Both orders give the same logits up to $10^{-4}$ in fp32 at every size (\texttt{scripts/check\_mix\_order.py}). Table~\ref{tab:main} and Table~\ref{tab:traincost} report the faster order per size and measure.

\begin{table*}[h]
\centering\footnotesize
\begin{tabular}{l r rrrrr r}
\toprule
 & \multicolumn{6}{c}{training: time per token relative to vanilla $\downarrow$} & prefill \\
\cmidrule(lr){2-7}\cmidrule(lr){8-8}
size & vanilla (tok/s) & DenseFormer & hyper-conn. & mHC & MUDDFormer & \dagf{} & vanilla (tok/s) \\
\midrule
75M & 108.8K & 1.02 & 1.36 & 1.63 & 1.17 & 1.44 & 851K \\
150M & 79.7K & 1.03 & 1.50 & 1.66 & 1.21 & 1.57 & 474K \\
300M & 49.1K & 1.03 & 1.59 & 1.68 & 1.28 & 1.77 & 245K \\
1B & 18.3K & 1.03 & 1.56 & 1.51 & 1.19 & 1.65 & 76K \\
\bottomrule
\end{tabular}

\caption{Measured training cost: time per token of a forward and backward pass relative to the vanilla Transformer of the same size (every model compiled, bf16 weights, 2 sequences of 1024 tokens, one A100-40GB, median of 20 iterations, fastest of two runs; \dagf{} in the faster of its two computation orders), with the vanilla Transformer's throughput in tokens per second for training and prefill. Prefill ratios are in Table~\ref{tab:main}.}
\label{tab:traincost}
\end{table*} The modular variant projects $2l{-}1$ sources per layer and costs 1.33 to 2.30$\times$.

\section{Held-out sets}
\label{app:cache}
The 21B-token tokenization's held-out tail is the last windows of its last shard, and the eight tokenization workers each took a contiguous block of Dolma's source-sorted file list, so that tail is templated instruction and Wikipedia text only; 21B-corpus runs score about 2.2 nats on it while scoring about 3.5 on WikiText-2. The tail is used only to compare runs within that corpus (the 300M three-way and the 1B pairs, where all models see the same training mix). The 1.7B-token slice used by the locality ablation has a similar single-source tail. All cross-corpus numbers in the paper are on WikiText-2, MathInstruct and GSM8K.

\section{Additional tables}
\label{app:tables}
This appendix gives the complete results that the body summarizes: every stage of the downstream and chat evaluations, the post-hoc baselines and learning rates, the training hyperparameters, every held-out set for the matched pairs, the routing-source ablation and the static substitution, the earlier pruning study, and the chat evaluation by category.

\subsection{Downstream and chat per stage}
\label{app:downstream}
Table~\ref{tab:downstream_tasks} gives the per-task accuracies of Table~\ref{tab:downstream}, Table~\ref{tab:downstream_full} every stage of it and Table~\ref{tab:chat_full} every chat comparison, including the second instruction-finetuning seed of the 300M pair. Repeating the 300M instruction finetuning with a second seed moves the 12-task mean by at most 0.001 and the routed-minus-vanilla gaps by at most 0.003. Alpaca finetuning leaves the multiple-choice mean within 0.002, SmolTalk lowers it by 0.7 to 1.4 points for every model, and GSM8K bits per byte improve by 0.13 to 0.25 after SmolTalk. Absolute judge scores are about 2 of 10 for every model; judging both orders cancels the judge's position bias (answer A wins 28 to 42\% of verdicts) at the cost of many ties.

\begin{table}[t]
\centering\footnotesize
\setlength{\tabcolsep}{2pt}
\begin{tabular}{llrrr}
\toprule
 & & MC acc. & \multicolumn{2}{c}{bits per byte $\downarrow$} \\
model & stage & $\uparrow$ & GSM8K & WikiText \\
\midrule
\multicolumn{5}{l}{\emph{300M, 21B corpus}} \\
vanilla & base & 0.369 & 1.458 & 1.128 \\
\dagf{} & base & \textbf{0.391} & \textbf{1.344} & \textbf{1.057} \\
vanilla & +Alpaca & 0.371 & 1.452 & 1.184 \\
\dagf{} & +Alpaca & \textbf{0.390} & \textbf{1.295} & \textbf{1.108} \\
vanilla & +SmolTalk & 0.364 & 1.269 & 1.309 \\
\dagf{} & +SmolTalk & \textbf{0.377} & \textbf{1.097} & \textbf{1.207} \\
\midrule
\multicolumn{5}{l}{\emph{1B, 21B corpus}} \\
vanilla 5B & base & 0.385 & 1.375 & 1.058 \\
vanilla 10B & base & 0.395 & 1.323 & 1.022 \\
\dagf{} 5B & base & 0.411 & 1.242 & 0.990 \\
\dagf{} 10B & base & \textbf{0.421} & \textbf{1.239} & \textbf{0.963} \\
vanilla 5B & +Alpaca & 0.385 & 1.348 & 1.090 \\
vanilla 10B & +Alpaca & 0.394 & 1.290 & 1.052 \\
\dagf{} 5B & +Alpaca & 0.410 & 1.200 & 1.021 \\
\dagf{} 10B & +Alpaca & \textbf{0.421} & \textbf{1.182} & \textbf{0.993} \\
vanilla 5B & +SmolTalk & 0.377 & 1.202 & 1.150 \\
vanilla 10B & +SmolTalk & 0.388 & 1.153 & 1.106 \\
\dagf{} 5B & +SmolTalk & 0.402 & 1.038 & 1.067 \\
\dagf{} 10B & +SmolTalk & \textbf{0.407} & \textbf{1.020} & \textbf{1.036} \\
\bottomrule
\end{tabular}
\caption{Downstream evaluation before (base) and after each instruction-finetuning recipe, summarized (Table~\ref{tab:downstream_tuned} gives per-task accuracy). MC acc.: mean accuracy over the twelve tasks (eleven multiple-choice and LAMBADA); GSM8K and WikiText: bits per byte on the gold GSM8K solutions and on WikiText (lower is better). One pretraining run per model. Repeating the 300M instruction finetuning with a second seed moves the 12-task mean by at most 0.001 and WikiText bpb by less than 0.001 (BoolQ alone by up to 0.02), and the routed-minus-vanilla gaps reproduce within 0.003, so the instruction-finetuning stage adds little noise; pretraining-seed variance is not measured. Shared-slice 75M and 150M pairs show the same pattern (Appendix~\ref{app:tables}).}
\label{tab:downstream_full}
\end{table}

\begin{table*}[t]
\centering\fontsize{8.1pt}{9.8pt}\selectfont
\setlength{\tabcolsep}{1.15pt}
\begin{tabular*}{\textwidth}{@{\extracolsep{\fill}}ll rrrrrrrrrrrrr @{\hspace{5pt}}|r}
\toprule
 &  & \multicolumn{14}{c}{accuracy (\%) $\uparrow$} \\
\cmidrule(lr){3-16}
size & model & ARC-e & ARC-c & HSwag & PIQA & WinoG & OBQA & SciQ & BoolQ & LAMB & CSQA & SIQA & MathQA & GSM8K & {\fontsize{9pt}{9.8pt}\selectfont \textbf{Mean}} \\
\midrule
300M & vanilla & 38.1 & 20.6 & 27.4 & 61.2 & 52.2 & 24.6 & 62.5 & 56.7 & 19.5 & \textbf{20.5} & 37.7 & 22.4 & 1.0 & {\fontsize{9pt}{9.8pt}\selectfont 34.2} \\
 & DenseFormer & 37.8 & 21.2 & 27.3 & 60.8 & 51.1 & 22.4 & 64.5 & 60.8 & 20.7 & 19.7 & 35.9 & 22.9 & 2.5 & {\fontsize{9pt}{9.8pt}\selectfont 34.4} \\
 & hyper-conn. & \textbf{39.6} & 22.5 & 27.8 & 60.8 & 48.6 & 25.8 & 65.9 & 55.0 & 21.6 & 19.7 & 35.8 & 21.6 & \textbf{3.0} & {\fontsize{9pt}{9.8pt}\selectfont 34.4} \\
 & mHC & 38.7 & \textbf{22.9} & 27.6 & 61.2 & 50.0 & \textbf{26.2} & 67.4 & \textbf{61.3} & 22.2 & 19.0 & \textbf{38.0} & \textbf{23.3} & 2.5 & {\fontsize{9pt}{9.8pt}\selectfont 35.4} \\
 & MUDDFormer & 38.3 & 22.6 & \textbf{28.5} & 60.8 & 52.2 & 25.0 & 68.4 & 58.7 & 23.0 & 19.7 & 37.3 & 22.0 & 1.0 & {\fontsize{9pt}{9.8pt}\selectfont 35.2} \\
 & \dagf{} & 38.8 & 22.1 & 28.2 & \textbf{61.7} & \textbf{52.3} & 24.4 & \textbf{73.7} & 60.4 & \textbf{26.5} & 19.6 & 37.8 & 23.1 & 2.5 & {\fontsize{9pt}{9.8pt}\selectfont \textbf{36.2}} \\
\midrule
1B (5B) & vanilla & 39.7 & 22.2 & 29.2 & 61.6 & \textbf{51.5} & 26.0 & 67.5 & 60.1 & 24.1 & 20.5 & 37.4 & 22.2 & 2.0 & {\fontsize{9pt}{9.8pt}\selectfont 35.7} \\
 & DenseFormer & 40.6 & 21.6 & 29.4 & 62.8 & 50.2 & 27.6 & 68.2 & \textbf{60.6} & 25.4 & 20.9 & 37.5 & 22.3 & 0.5 & {\fontsize{9pt}{9.8pt}\selectfont 36.0} \\
 & hyper-conn. & 42.1 & 22.7 & 29.9 & \textbf{64.7} & \textbf{51.5} & 25.0 & 72.8 & 57.2 & 26.1 & 19.5 & 37.8 & 22.0 & 2.0 & {\fontsize{9pt}{9.8pt}\selectfont 36.4} \\
 & mHC & 41.4 & 23.8 & 31.0 & 64.5 & 51.2 & \textbf{27.8} & 75.6 & 59.9 & 27.7 & 20.1 & 38.4 & 22.5 & \textbf{2.5} & {\fontsize{9pt}{9.8pt}\selectfont 37.4} \\
 & MUDDFormer & \textbf{42.9} & 23.2 & 30.2 & 63.0 & 50.9 & 27.4 & 76.3 & 49.3 & 27.2 & 20.5 & 39.0 & \textbf{23.4} & 1.0 & {\fontsize{9pt}{9.8pt}\selectfont 36.5} \\
 & \dagf{} & 42.1 & \textbf{24.0} & \textbf{32.1} & 63.7 & 51.0 & 26.6 & \textbf{79.3} & 58.9 & \textbf{30.5} & \textbf{22.9} & \textbf{39.3} & 22.8 & 2.0 & {\fontsize{9pt}{9.8pt}\selectfont \textbf{38.1}} \\
\midrule
1B (10B) & vanilla & 41.5 & 23.5 & 30.5 & 63.8 & 50.6 & 26.4 & 72.6 & 59.5 & 26.4 & 19.0 & 38.4 & 22.0 & 1.0 & {\fontsize{9pt}{9.8pt}\selectfont 36.6} \\
 & \dagf{} & \textbf{43.4} & \textbf{24.2} & \textbf{33.7} & \textbf{65.1} & \textbf{51.9} & \textbf{27.8} & \textbf{81.0} & \textbf{60.6} & \textbf{33.3} & \textbf{21.5} & \textbf{39.6} & \textbf{23.1} & \textbf{2.5} & {\fontsize{9pt}{9.8pt}\selectfont \textbf{39.1}} \\
\bottomrule
\end{tabular*}

\caption{Per-task accuracy (\%) of the base models in Table~\ref{tab:downstream}, normalized for ARC, HellaSwag, PIQA, OpenBookQA and MathQA, GSM8K 3-shot exact match with flexible numeric extraction on the first 200 test problems; Mean: over the 13 tasks. Between the vanilla Transformer and \dagf{}: the baselines of Table~\ref{tab:main}. Bold: best per block. 1B (5B), 1B (10B): the 1B models pretrained on 5B and 10B tokens.}
\label{tab:downstream_tasks}
\end{table*}

\begin{table*}[t]
\centering\fontsize{8.1pt}{9.8pt}\selectfont
\setlength{\tabcolsep}{1.15pt}
\begin{tabular*}{\textwidth}{@{\extracolsep{\fill}}lll rrrrrrrrrrrrr r}
\toprule
 &  &  & \multicolumn{14}{c}{accuracy (\%) $\uparrow$} \\
\cmidrule(lr){4-17}
size & stage & model & ARC-e & ARC-c & HSwag & PIQA & WinoG & OBQA & SciQ & BoolQ & LAMB & CSQA & SIQA & MathQA & GSM8K & \textbf{Mean} \\
\midrule
300M & base & vanilla & 38.1 & 20.6 & 27.4 & 61.2 & 52.2 & 24.6 & 62.5 & 56.7 & 19.5 & \textbf{20.5} & 37.7 & 22.4 & 1.0 & 34.2 \\
 &  & DenseFormer & 37.8 & 21.2 & 27.3 & 60.8 & 51.1 & 22.4 & 64.5 & \textbf{60.8} & 20.7 & 19.7 & 35.9 & 22.9 & \textbf{2.5} & 34.4 \\
 &  & MUDDFormer & 38.3 & \textbf{22.6} & \textbf{28.5} & 60.8 & 52.2 & \textbf{25.0} & 68.4 & 58.7 & 23.0 & 19.7 & 37.3 & 22.0 & 1.0 & 35.2 \\
 &  & \dagf{} & \textbf{38.8} & 22.1 & 28.2 & \textbf{61.7} & \textbf{52.3} & 24.4 & \textbf{73.7} & 60.4 & \textbf{26.5} & 19.6 & \textbf{37.8} & \textbf{23.1} & \textbf{2.5} & \textbf{36.2} \\
\addlinespace[1pt]
 & +Alpaca & vanilla & 37.1 & 21.8 & 27.7 & 60.6 & \textbf{53.7} & 25.6 & 59.3 & 59.1 & 20.8 & 20.0 & 37.4 & \textbf{22.6} & \textbf{2.5} & 34.5 \\
 &  & DenseFormer & 38.5 & \textbf{23.6} & 27.4 & 60.9 & 52.1 & 25.6 & 60.2 & \textbf{62.0} & 21.9 & \textbf{20.6} & 36.9 & 22.1 & 0.5 & 34.8 \\
 &  & MUDDFormer & \textbf{39.6} & \textbf{23.6} & 28.6 & 60.1 & 51.7 & 26.4 & \textbf{67.1} & 59.3 & 24.6 & 20.1 & 38.1 & 21.7 & 0.5 & 35.5 \\
 &  & \dagf{} & 38.0 & 23.4 & \textbf{28.7} & \textbf{61.9} & 52.4 & \textbf{27.2} & 65.6 & 61.8 & \textbf{28.7} & 19.5 & \textbf{38.7} & \textbf{22.6} & 2.0 & \textbf{36.2} \\
\addlinespace[1pt]
 & +SmolTalk & vanilla & 36.1 & 22.3 & 27.6 & 58.3 & 51.0 & \textbf{27.6} & 57.1 & \textbf{61.9} & 14.2 & 19.6 & 37.2 & 23.5 & \textbf{2.0} & 33.7 \\
 &  & DenseFormer & 37.0 & 21.3 & 27.8 & \textbf{59.9} & 51.1 & 25.2 & 59.6 & 59.3 & 15.5 & \textbf{20.1} & 35.6 & 23.6 & 0.5 & 33.6 \\
 &  & MUDDFormer & 37.1 & 22.5 & 28.8 & 58.8 & 48.9 & 26.8 & 63.6 & 60.5 & 17.1 & 19.8 & 37.3 & \textbf{24.9} & 1.5 & 34.4 \\
 &  & \dagf{} & \textbf{37.5} & \textbf{23.1} & \textbf{29.0} & 59.8 & \textbf{52.6} & \textbf{27.6} & \textbf{68.2} & 52.4 & \textbf{22.3} & 19.7 & \textbf{37.5} & 23.2 & 0.5 & \textbf{34.9} \\
\midrule
1B & base & vanilla & 39.7 & 22.2 & 29.2 & 61.6 & \textbf{51.5} & 26.0 & 67.5 & 60.1 & 24.1 & 20.5 & 37.4 & 22.2 & 2.0 & 35.7 \\
 &  & DenseFormer & 40.6 & 21.6 & 29.4 & 62.8 & 50.2 & 27.6 & 68.2 & \textbf{60.6} & 25.4 & 20.9 & 37.5 & 22.3 & 0.5 & 36.0 \\
 &  & mHC & 41.4 & 23.8 & 31.0 & \textbf{64.5} & 51.2 & \textbf{27.8} & 75.6 & 59.9 & 27.7 & 20.1 & 38.4 & 22.5 & \textbf{2.5} & 37.4 \\
 &  & MUDDFormer & \textbf{42.9} & 23.2 & 30.2 & 63.0 & 50.9 & 27.4 & 76.3 & 49.3 & 27.2 & 20.5 & 39.0 & \textbf{23.4} & 1.0 & 36.5 \\
 &  & \dagf{} & 42.1 & \textbf{24.0} & \textbf{32.1} & 63.7 & 51.0 & 26.6 & \textbf{79.3} & 58.9 & \textbf{30.5} & \textbf{22.9} & \textbf{39.3} & 22.8 & 2.0 & \textbf{38.1} \\
\addlinespace[1pt]
(5B) & +Alpaca & vanilla & 40.2 & 24.1 & 29.4 & 61.9 & 50.7 & 27.0 & 63.1 & \textbf{61.8} & 25.5 & 18.5 & 38.1 & 21.9 & 1.5 & 35.7 \\
 &  & DenseFormer & 40.4 & 23.2 & 29.7 & 61.9 & 50.1 & 28.0 & 64.8 & 59.3 & 27.5 & 19.2 & 38.2 & 22.7 & \textbf{2.0} & 35.9 \\
 &  & mHC & 41.7 & 23.7 & 31.2 & 63.8 & 50.2 & 27.6 & 68.0 & 60.7 & 28.9 & 20.6 & 38.4 & 21.7 & \textbf{2.0} & 36.8 \\
 &  & MUDDFormer & 41.7 & \textbf{25.0} & 30.7 & \textbf{63.9} & \textbf{50.9} & 28.6 & 72.1 & 59.6 & 29.5 & 19.6 & 39.1 & \textbf{24.2} & 1.5 & 37.4 \\
 &  & \dagf{} & \textbf{42.0} & 24.1 & \textbf{32.2} & 63.3 & 50.6 & \textbf{29.4} & \textbf{76.4} & 56.0 & \textbf{33.8} & \textbf{22.4} & \textbf{40.0} & 22.4 & 1.5 & \textbf{38.0} \\
\addlinespace[1pt]
 & +SmolTalk & vanilla & 38.2 & 22.5 & 29.3 & 62.1 & 50.4 & 26.6 & 60.4 & \textbf{62.1} & 18.8 & \textbf{20.8} & 37.8 & 23.7 & 1.0 & 34.9 \\
 &  & DenseFormer & 38.4 & 22.2 & 29.6 & 61.7 & 50.2 & 27.2 & 64.8 & 61.2 & 20.1 & 18.8 & 37.1 & 23.3 & 1.5 & 35.1 \\
 &  & mHC & 40.7 & 24.5 & 30.7 & 62.8 & 50.7 & 26.4 & 66.0 & 61.5 & 23.5 & 20.2 & 37.8 & \textbf{23.9} & \textbf{2.0} & 36.2 \\
 &  & MUDDFormer & \textbf{41.7} & \textbf{24.6} & 29.9 & 62.6 & \textbf{52.1} & \textbf{27.8} & 70.2 & 56.1 & 23.4 & 19.1 & \textbf{38.6} & 23.0 & 1.5 & 36.2 \\
 &  & \dagf{} & 41.1 & 22.8 & \textbf{32.0} & \textbf{63.4} & 50.6 & \textbf{27.8} & \textbf{75.2} & 61.1 & \textbf{27.1} & 20.6 & 38.0 & 23.0 & 1.5 & \textbf{37.2} \\
\midrule
1B & base & vanilla & 41.5 & 23.5 & 30.5 & 63.8 & 50.6 & 26.4 & 72.6 & 59.5 & 26.4 & 19.0 & 38.4 & 22.0 & 1.0 & 36.6 \\
 &  & \dagf{} & \textbf{43.4} & \textbf{24.2} & \textbf{33.7} & \textbf{65.1} & \textbf{51.9} & \textbf{27.8} & \textbf{81.0} & \textbf{60.6} & \textbf{33.3} & \textbf{21.5} & \textbf{39.6} & \textbf{23.1} & \textbf{2.5} & \textbf{39.1} \\
\addlinespace[1pt]
(10B) & +Alpaca & vanilla & 40.6 & 23.9 & 30.5 & 62.8 & 51.0 & 27.0 & 67.3 & \textbf{61.8} & 27.8 & 20.1 & 38.2 & 22.2 & \textbf{4.5} & 36.7 \\
 &  & \dagf{} & \textbf{43.4} & \textbf{25.5} & \textbf{34.1} & \textbf{65.6} & \textbf{51.3} & \textbf{28.6} & \textbf{75.3} & 60.3 & \textbf{37.4} & \textbf{22.2} & \textbf{38.8} & \textbf{22.5} & 1.0 & \textbf{38.9} \\
\addlinespace[1pt]
 & +SmolTalk & vanilla & 40.0 & 22.4 & 30.6 & 62.5 & 50.0 & 27.6 & 67.0 & 59.6 & 21.4 & \textbf{21.5} & 38.5 & \textbf{23.9} & \textbf{3.5} & 36.0 \\
 &  & \dagf{} & \textbf{41.8} & \textbf{23.3} & \textbf{33.0} & \textbf{64.3} & \textbf{51.1} & \textbf{28.0} & \textbf{75.4} & \textbf{61.1} & \textbf{29.9} & 17.9 & \textbf{39.0} & 23.3 & 1.5 & \textbf{37.7} \\
\bottomrule
\end{tabular*}

\caption{The evaluation tasks of Table~\ref{tab:downstream} at every stage: the base models and the same models after each instruction-finetuning recipe. The 13-task mean falls after SmolTalk for every model, vanilla and routed alike, while the same models fit the instruction data better (Table~\ref{tab:instruct}) and their bits per byte on the gold GSM8K solutions improve (1.344 to 1.097 for the 300M \dagf{}, 1.242 to 1.038 at 1B): instruction finetuning moves the output distribution away from zero-shot multiple choice, and the loss concentrates in LAMBADA, SciQ and BoolQ. This is why the body compares instruction-finetuned models on instruction following instead. Bold: best per stage.}
\label{tab:downstream_tuned}
\end{table*}

\begin{table}[t]
\centering\footnotesize
\setlength{\tabcolsep}{2.2pt}
\begin{tabular}{llrc}
\toprule
SFT & \dagf{} vs.\ vanilla & win rate $\uparrow$ [CI] & W/L \\
\midrule
Alpaca & 300M vs.\ 300M, seed 1 & 0.57 [0.50, 0.64] & 16/10 \\
Alpaca & 300M vs.\ 300M, seed 2 & 0.61 [0.54, 0.67] & 19/8 \\
Alpaca & 1B@5B vs.\ 1B@5B & 0.58 [0.51, 0.66] & 26/12 \\
Alpaca & 1B@5B vs.\ 1B@10B & 0.56 [0.48, 0.64] & 23/17 \\
Alpaca & 1B@10B vs.\ 1B@10B & 0.56 [0.48, 0.63] & 19/14 \\
SmolTalk & 300M vs.\ 300M, seed 1 & \textbf{0.69} [0.63, 0.75] & 27/2 \\
SmolTalk & 300M vs.\ 300M, seed 2 & 0.64 [0.57, 0.71] & 30/5 \\
SmolTalk & 1B@5B vs.\ 1B@5B & 0.59 [0.52, 0.66] & 23/8 \\
SmolTalk & 1B@5B vs.\ 1B@10B & 0.60 [0.52, 0.67] & 27/10 \\
SmolTalk & 1B@10B vs.\ 1B@10B & 0.64 [0.57, 0.72] & 30/9 \\
\bottomrule
\end{tabular}
\caption{Judged chat quality on the 80 MT-Bench first turns. Win rate of \dagf{} against the vanilla Transformer, judged by Qwen3-Coder-30B-A3B-Instruct \citep{qwen3,qwen3coder} with the MT-Bench pairwise prompts (with the reference answer for math, reasoning and coding), with ties counted as half, averaged over both answer orders, with a 95\% bootstrap interval; W/L counts prompts where both orders agree. ``Seed 1/2'' are two independent instruction-finetuning runs of both models from the same pretrained checkpoints. The judge prefers \dagf{} in every comparison; the interval excludes 0.5 in seven of ten.}
\label{tab:chat_full}
\end{table}

\begin{table*}[t]
\centering\small
\setlength{\tabcolsep}{4pt}
\begin{tabular}{ll cc cc}
\toprule
 & & \multicolumn{2}{c}{+Alpaca} & \multicolumn{2}{c}{+SmolTalk} \\
\cmidrule(lr){3-4}\cmidrule(lr){5-6}
size & routed model & win rate & W/T/L & win rate & W/T/L \\
\midrule
300M & \dagf{} & 0.65$^\dagger$ [0.60, 0.70] & 124/69/55 & 0.70$^\dagger$ [0.65, 0.75] & 138/69/41 \\
 & MUDDFormer & 0.63$^\dagger$ [0.58, 0.68] & 117/78/53 & 0.64$^\dagger$ [0.59, 0.69] & 120/73/55 \\
 & DenseFormer & 0.55$^\dagger$ [0.50, 0.61] & 91/84/73 & 0.57$^\dagger$ [0.52, 0.63] & 101/81/66 \\
 & hyper-conn. & 0.59$^\dagger$ [0.54, 0.64] & 109/80/60 & 0.58$^\dagger$ [0.53, 0.64] & 106/76/66 \\
 & mHC & 0.58$^\dagger$ [0.53, 0.63] & 100/89/60 & 0.62$^\dagger$ [0.57, 0.67] & 118/73/57 \\
\midrule
1B (5B) & \dagf{} & 0.63$^\dagger$ [0.58, 0.68] & 120/71/56 & 0.66$^\dagger$ [0.61, 0.71] & 129/66/53 \\
 & MUDDFormer & 0.57$^\dagger$ [0.52, 0.63] & 101/79/68 & 0.56$^\dagger$ [0.50, 0.61] & 104/71/73 \\
 & DenseFormer & 0.46 [0.41, 0.52] & 75/81/91 & 0.50 [0.45, 0.56] & 86/80/82 \\
 & mHC & 0.55 [0.49, 0.60] & 94/77/76 & 0.61$^\dagger$ [0.56, 0.66] & 120/65/64 \\
 & hyper-conn. & 0.51 [0.46, 0.57] & 85/82/80 & 0.56$^\dagger$ [0.51, 0.61] & 97/79/72 \\
\midrule
1B (10B) & \dagf{} & 0.66$^\dagger$ [0.61, 0.71] & 122/79/47 & 0.70$^\dagger$ [0.65, 0.75] & 143/57/48 \\
\midrule
equal compute & \dagf{} 5B vs vanilla 10B & 0.60$^\dagger$ [0.55, 0.65] & 107/75/66 & 0.58$^\dagger$ [0.53, 0.63] & 114/61/73 \\
\bottomrule
\end{tabular}

\caption{Every judged AlpacaEval comparison, with the intervals and counts behind the win rates of Table~\ref{tab:instruct}: AlpacaEval \citep{alpaca_eval} win rate of each routed model against the vanilla Transformer of the same size and recipe, on a stratified 249-prompt subset of the 805 (sampling seed 1234). Every pair is judged in both answer orders by \texttt{gpt-5.6-sol} (2026-10-09); the win rate counts wins plus half the ties, averaged over the two orders and then over prompts, with a 95\% bootstrap interval. W/T/L counts a prompt as a win or a loss only when both orders agree, so disagreements become ties. $\dagger$: the interval excludes 0.5. Last row: \dagf{} trained on 5B tokens against the vanilla Transformer trained on 10B, at about equal compute.}
\label{tab:alpaca}
\end{table*}

\subsection{Post-hoc baselines and learning rates}
\label{app:posthoc}

The hyper-connection baselines use one connection module per attention and MLP sub-layer, as in their papers. Hyper-connections have no official code release (the paper gives PyTorch-style pseudocode), so we use a widely used open implementation\footnote{\url{https://github.com/lucidrains/hyper-connections}} with the paper's LayerNorm and initialization; mHC uses DeepSeek's official PyTorch reference operations\footnote{\url{https://github.com/deepseek-ai/TileKernels}} with the paper's settings (4 streams, gating factors initialized to 0.01, 20 Sinkhorn-Knopp iterations). Alpaca is used in its cleaned release\footnote{\url{https://huggingface.co/datasets/yahma/alpaca-cleaned}}.

\begin{table*}[t]
\centering\footnotesize
\setlength{\tabcolsep}{3pt}
\begin{tabular}{llrrrr}
\toprule
run & size & WikiText-2 & MathInstruct & GSM8K & balanced \\
\midrule
vanilla & 75M & 4.997 & 4.271 & 4.242 & 5.167 \\
vanilla, hidden 656 & 75M & 4.769 & 4.057 & 4.052 & 4.961 \\
vanilla, 13 layers & 75M & 4.871 & 4.215 & 4.174 & 5.094 \\
DenseFormer & 75M & 5.027 & 4.303 & 4.249 & 5.196 \\
hyper-connections (static) & 75M & 5.042 & 4.301 & 4.246 & 5.191 \\
hyper-connections (DHC$\times$4) & 75M & 4.964 & 4.202 & 4.157 & 5.114 \\
mHC & 75M & 4.913 & 4.196 & 4.143 & 5.106 \\
MUDDFormer & 75M & 4.974 & 4.269 & 4.130 & 5.166 \\
\dagf{} & 75M & \textbf{4.711} & \textbf{3.984} & \textbf{3.973} & \textbf{4.937} \\
\midrule
vanilla & 150M & 4.268 & 3.599 & 3.608 & 4.471 \\
vanilla, hidden 864 & 150M & 4.199 & 3.537 & 3.599 & 4.409 \\
DenseFormer & 150M & 4.246 & 3.581 & 3.579 & 4.460 \\
hyper-connections (static) & 150M & 4.247 & 3.595 & 3.587 & 4.471 \\
hyper-connections (DHC$\times$4) & 150M & 4.220 & 3.546 & 3.620 & 4.426 \\
mHC & 150M & 4.209 & 3.535 & 3.568 & 4.421 \\
MUDDFormer & 150M & \textbf{4.041} & \textbf{3.377} & \textbf{3.357} & \textbf{4.269} \\
\dagf{} & 150M & 4.085 & 3.383 & 3.402 & 4.305 \\
\midrule
vanilla & 300M & 3.808 & 2.742 & 3.232 & 3.832 \\
DenseFormer & 300M & 3.745 & 2.714 & 3.249 & 3.782 \\
hyper-connections (DHC$\times$4) & 300M & 3.692 & 2.676 & 3.157 & 3.734 \\
mHC & 300M & 3.652 & 2.639 & 3.137 & 3.706 \\
MUDDFormer & 300M & 3.649 & 2.648 & 3.032 & 3.703 \\
\dagf{} & 300M & \textbf{3.544} & \textbf{2.565} & \textbf{2.937} & \textbf{3.606} \\
\midrule
vanilla & 1B, 5B tok. & 3.569 & 2.586 & 3.058 & 3.617 \\
DenseFormer & 1B, 5B tok. & 3.513 & 2.563 & 3.006 & 3.587 \\
hyper-connections (DHC$\times$4) & 1B, 5B tok. & 3.492 & 2.544 & 3.031 & 3.551 \\
mHC & 1B, 5B tok. & 3.438 & 2.508 & 2.947 & 3.506 \\
MUDDFormer & 1B, 5B tok. & 3.423 & 2.499 & 2.897 & 3.501 \\
\dagf{} & 1B, 5B tok. & \textbf{3.307} & \textbf{2.403} & \textbf{2.755} & \textbf{3.404} \\
\bottomrule
\end{tabular}
\caption{Post-hoc baselines of Table~\ref{tab:main} on every common held-out set (NLL, nats per token). The parameter-matched vanilla Transformers carry \dagf{}'s routing parameters as extra width (hidden 656 and 864; 107M and 182M parameters) or, at 75M, as extra depth (13 layers, 106M); hyper-connections are the static 4-stream variant. All runs of a size share recipe, data, tokens and seed; the 75M and 150M runs use the 12B' corpus and the 300M and 1B runs the 21B corpus.}
\label{tab:posthoc_sets}
\end{table*}

\begin{table}[t]
\centering\footnotesize
\setlength{\tabcolsep}{3pt}
\begin{tabular}{lrrrr}
\toprule
75M, WikiText-2 NLL $\downarrow$ & 2.5e-4 & 5e-4 & 1e-3 & 2e-3 \\
\midrule
vanilla & 5.878 & 4.997 & 4.654 & 4.563 \\
\dagf{} & 5.558 & 4.711 & 4.413 & 4.329 \\
gap & 0.32 & 0.29 & 0.24 & 0.23 \\
\bottomrule
\end{tabular}
\caption{Learning-rate sweep at 75M (12B' corpus, 1.58B tokens; the predictor's learning rate is 0.6$\times$ the backbone's). Both models improve up to the largest rate; the matched-rate gap moves from 0.32 to 0.23 nats and the best-against-best gap is 0.23. The same holds on MathInstruct, GSM8K and the balanced set (best-against-best narrows the 5e-4 gap by 5 to 19\%; 30\% on the Dolma tail).}
\label{tab:lrsweep}
\end{table}

\begin{table}[h]
\centering\scriptsize
\setlength{\tabcolsep}{2.4pt}
\begin{tabular}{lrrrrrr}
\toprule
routing source (75M) & routing & \multicolumn{4}{c}{held-out NLL $\downarrow$} & tasks \\
\cmidrule(lr){3-6}
 & params & own & WikiText & Math & GSM8K & acc.\ $\uparrow$ \\
\midrule
none (vanilla) & 0 & 4.213 & 6.593 & 5.373 & 5.404 & 25.7 \\
static table, no input & 500 & 4.171 & 6.529 & 5.243 & 5.279 & n/a \\
per-layer predictors & 35.6M & 4.058 & 6.256 & 5.025 & 5.148 & 26.0 \\
predictor only & 28.7M & 4.031 & 6.207 & 4.981 & 5.116 & 26.2 \\
predictor + refinement & 29.1M & \textbf{4.014} & \textbf{6.163} & \textbf{4.932} & \textbf{5.080} & \textbf{26.5} \\
\bottomrule
\end{tabular}
\caption{Routing-source ablation: one 75M backbone, granularity and 1.57B tokens of a 1.7B-token Dolma slice, varying only the source of $\valpha$. Static table: one learned weight per read and source, no input; per-layer predictors: an encoder over the token ids per layer, nothing shared; predictor + refinement is \dagf{}. Own: the run's held-out Dolma tail; Math: MathInstruct; tasks acc.: the 13-task mean of Table~\ref{tab:downstream} before instruction finetuning (the static table was not run on the tasks).}
\label{tab:locality}
\end{table}

\subsection{Training hyperparameters}
\label{app:hparams}
Table~\ref{tab:hparams} lists the architecture, token budget and optimization of every pretraining run and the instruction-tuning recipe. The routed baselines use the settings of the pair of their size (Section~\ref{sec:expsetup}).
\begin{table*}[h]\centering\footnotesize\setlength{\tabcolsep}{3pt}
\begin{tabular}{lcccc}\toprule
size & 75M & 150M & 300M & 1B \\\midrule
layers / width / heads & 6 / 512 / 8 & 8 / 768 / 12 & 12 / 1024 / 16 & 16 / 2048 / 16 \\
MLP width & 2048 & 3072 & 4096 & 8192 \\
backbone params (M) & 76.6 & 152.6 & 304.1 & 1279.4 \\
routing params (M) & 29.1 & 30.0 & 32.3 & 36.7 \\
steps & 3000 & 6000 & 12000 & 9540 (+9540) \\
tokens (B) & 1.57 & 3.15 & 6.29 & 5.0 (10.0) \\
peak LR backbone / predictor & $5\mathrm{e}{-4}$ / $3\mathrm{e}{-4}$ & $5\mathrm{e}{-4}$ / $3\mathrm{e}{-4}$ & $5\mathrm{e}{-4}$ / $3\mathrm{e}{-4}$ & $4\mathrm{e}{-4}$ / $2\mathrm{e}{-4}$ \\
warmup steps & 300 & 300 & 300 & 500 \\
\dagf{} V normalization (RMSNorm on mixed $v_h$) & no & no & yes & yes \\
routed FLOPs / vanilla & 1.18 & 1.33 & 1.65 & 2.13 \\
\bottomrule\end{tabular}
\caption{Pretraining configurations. Common to all: 1024-token sequences, 512 sequences (524K tokens) per step, AdamW $(0.9, 0.95)$, weight decay 0.1, gradient clipping 1.0, linear decay to zero, bf16, tied embeddings, vocabulary 100,352 (the OLMo-2 tokenizer). Predictor: 256-d token and position embeddings, 2 causal encoder layers with 4 heads, trunk width 512, local refinement width 128. The 1B runs were continued from 5B to 10B tokens with the schedule re-stretched and the learning rate re-warmed over 500 steps. Instruction tuning: global batch 64 sequences of 1024 tokens, cosine learning rate $2{\times}10^{-4}$ ($10^{-4}$ at 1B), assistant-only loss, ChatML format. The 1B hyper-connections model was pretrained and instruction-tuned on four H100 GPUs, with 8 sequences per micro-batch in pretraining against 2 for the other 1B models on A100s; the 1B mHC and DenseFormer instruction tuning used 2 sequences per micro-batch instead of 4 to fit in memory. Micro-batching changes neither the global batch nor the sequences in each step.}
\label{tab:hparams}\end{table*}

\subsection{Matched pairs on all held-out sets}
\label{app:pairs}
Table~\ref{tab:pairs_all} gives every matched pair on the four held-out sets of Section~\ref{sec:data}, including the 75M and 150M pairs on both tokenizations of the 12B-token slice and the 300M pair repeated with a second pretraining seed. \dagf{} has the lower loss in every cell. The second seed changes a model's loss by up to 0.08 nats (the vanilla Transformer on GSM8K) and by at most 0.03 on WikiText-2.
\begin{table*}[h]\centering\footnotesize\setlength{\tabcolsep}{2.5pt}
\begin{tabular}{llcccccccc}\toprule
size & corpus & \multicolumn{2}{c}{WikiText-2} & \multicolumn{2}{c}{MathInstruct} & \multicolumn{2}{c}{GSM8K} & \multicolumn{2}{c}{balanced Dolma} \\
 & & vanilla & \dagf{} & vanilla & \dagf{} & vanilla & \dagf{} & vanilla & \dagf{} \\\midrule
75M & 12B & 4.980 & 4.775 & 4.267 & 4.053 & 4.216 & 3.998 & 5.175 & 4.992 \\
75M & 12B' & 4.997 & 4.711 & 4.271 & 3.984 & 4.242 & 3.973 & 5.167 & 4.937 \\
150M & 12B & 4.297 & 4.100 & 3.630 & 3.424 & 3.626 & 3.465 & 4.514 & 4.334 \\
150M & 12B' & 4.268 & 4.085 & 3.599 & 3.383 & 3.608 & 3.402 & 4.471 & 4.305 \\
300M & 21B & 3.808 & 3.544 & 2.742 & 2.565 & 3.232 & 2.937 & 3.832 & 3.606 \\
300M, seed 2 & 21B & 3.835 & 3.530 & 2.774 & 2.562 & 3.308 & 2.897 & 3.855 & 3.608 \\
1B, 5B tok & 21B & 3.569 & 3.307 & 2.586 & 2.403 & 3.058 & 2.755 & 3.617 & 3.404 \\
1B, 10B tok & 21B & 3.446 & 3.219 & 2.483 & 2.321 & 2.900 & 2.644 & 3.501 & 3.314 \\
\bottomrule\end{tabular}
\caption{Held-out NLL (nats per token) of every matched pair on the four cross-corpus sets (final checkpoints): WikiText-2, MathInstruct, GSM8K and a source-balanced Dolma set (64 windows from each of 14 Dolma sources, drawn from files that no training corpus consumed; books, Wikipedia and open-web-math could not be represented). \dagf{} is lower in every cell; ``seed 2'' is the 300M pair repeated with a second pretraining seed.}
\label{tab:pairs_all}\end{table*}

\subsection{Gap along training}
\label{app:gapcurve}
The vanilla-minus-\dagf{} gap on each run's own held-out cache at matched steps (nats): 
75M (12B'): +0.096 @ 0.26B, +0.178 @ 0.53B, +0.175 @ 0.79B, +0.141 @ 1.32B; 150M (12B'): +0.160 @ 0.53B, +0.114 @ 1.32B, +0.107 @ 2.11B, +0.103 @ 2.91B; 300M (21B): +0.343 @ 1.31B, +0.207 @ 2.88B, +0.167 @ 4.46B, +0.161 @ 6.03B; 1B (21B): +0.197 @ 2.10B, +0.133 @ 4.72B, +0.116 @ 7.34B, +0.110 @ 9.96B. The gap opens within the first quarter of each run, peaks, and narrows slowly; the 1B entry is the 5B-token run.

\subsection{Router-locality ablation on all metrics}
\label{app:locality}
Table~\ref{tab:locality_all} extends the routing-source ablation of Section~\ref{sec:source} to every held-out set and to the multiple-choice mean before and after instruction tuning. The four-stream arms keep the same order on every set: the shared predictor with the local refinement (\dagf{}), the shared predictor alone, per-layer predictors, the static table and the vanilla Transformer. At 75M the multiple-choice means of all arms lie within 0.02 of each other.
\begin{table*}[h]\centering\footnotesize\setlength{\tabcolsep}{2.5pt}
\begin{tabular}{lccccc}\toprule
routing source & own cache & WikiText-2 & MathInstruct & GSM8K & MC w/o BoolQ \\
 & & & & & base / Alpaca / SmolTalk \\\midrule
none (vanilla) & 4.213 & 6.593 & 5.373 & 5.404 & 0.269 / 0.278 / 0.293 \\
static learned table (no token input) & 4.176 & 6.529 & 5.243 & 5.279 & - \\
per-layer predictors (unshared) & 4.058 & 6.256 & 5.025 & 5.148 & 0.271 / 0.290 / 0.304 \\
shared predictor only & 4.031 & 6.207 & 4.981 & 5.116 & 0.274 / 0.289 / 0.301 \\
shared + local (\dagf{}) & 4.014 & 6.163 & 4.932 & 5.080 & 0.276 / 0.290 / 0.302 \\
modular variant & 3.926 & 6.068 & 4.861 & 4.989 & 0.281 / 0.295 / 0.300 \\
modular + sparsity penalty & 4.091 & 6.274 & 5.139 & 5.229 & 0.273 / 0.286 / 0.300 \\
\bottomrule\end{tabular}
\caption{The 75M router-locality arms (1.57B tokens of a 1.7B-token Dolma slice, 4 GPUs each) on every held-out set, and the 11-task multiple-choice mean (without BoolQ, whose majority-class flips move the 12-task mean by up to 20 points at this size) before and after instruction tuning. The modular variant, which also routes the intra-layer edge, is best on loss at 75M; its routing cost is 1.33$\times$ the vanilla Transformer here and it is not better than FourWay routing at 150M or 300M (Appendix~\ref{app:modular}).}
\label{tab:locality_all}\end{table*}

\subsection{Shared-slice 75M, 150M and 300M pairs downstream}
\label{app:shared_down}
Table~\ref{tab:shared_down} evaluates the pairs of the earlier pruning study (Table~\ref{tab:pruning_shared}) on the evaluation tasks before and after instruction tuning. At every size and stage \dagf{} has the higher multiple-choice mean and the lower bits per byte on GSM8K solutions and WikiText.
\begin{table*}[h]\centering\footnotesize\setlength{\tabcolsep}{2.5pt}
\begin{tabular}{llccccc}\toprule
size & model & stage & MC acc. (12) & GSM8K acc. & GSM8K bpb & WikiText bpb \\\midrule
75M & vanilla & base & 0.304 & 0.010 & 1.840 & 1.393 \\
75M & vanilla & +Alpaca & 0.309 & 0.010 & 1.801 & 1.459 \\
75M & vanilla & +SmolTalk & 0.309 & 0.015 & 1.613 & 1.595 \\
75M & \dagf{} & base & 0.316 & 0.025 & 1.759 & 1.347 \\
75M & \dagf{} & +Alpaca & 0.319 & 0.015 & 1.733 & 1.417 \\
75M & \dagf{} & +SmolTalk & 0.336 & 0.010 & 1.502 & 1.550 \\
150M & vanilla & base & 0.336 & 0.010 & 1.593 & 1.196 \\
150M & vanilla & +Alpaca & 0.337 & 0.015 & 1.560 & 1.260 \\
150M & vanilla & +SmolTalk & 0.340 & 0.020 & 1.337 & 1.399 \\
150M & \dagf{} & base & 0.361 & 0.005 & 1.519 & 1.152 \\
150M & \dagf{} & +Alpaca & 0.366 & 0.005 & 1.491 & 1.211 \\
150M & \dagf{} & +SmolTalk & 0.361 & 0.010 & 1.219 & 1.333 \\
300M & vanilla & base & 0.379 & 0.025 & 1.402 & 1.065 \\
300M & vanilla & +Alpaca & 0.367 & 0.025 & 1.403 & 1.117 \\
300M & vanilla & +SmolTalk & 0.368 & 0.015 & 1.145 & 1.232 \\
300M & \dagf{} & base & 0.397 & 0.015 & 1.331 & 1.022 \\
300M & \dagf{} & +Alpaca & 0.395 & 0.015 & 1.324 & 1.072 \\
300M & \dagf{} & +SmolTalk & 0.393 & 0.010 & 1.055 & 1.165 \\
\bottomrule\end{tabular}
\caption{The original 12B-slice pairs before and after instruction tuning (the 300M \dagf{} of this slice stopped at 9000 of 12000 steps, so its margin is a lower bound). MC acc.: mean accuracy over the 12 multiple-choice tasks; GSM8K acc.: 3-shot exact match on the first 200 test problems; bpb: bits per byte of the GSM8K solutions and of WikiText.}
\label{tab:shared_down}\end{table*}

\subsection{Pruning: absolute losses, controls and immediate damage}
\label{app:prune}
Table~\ref{tab:prune_all} gives the absolute losses behind the earlier pruning study (Table~\ref{tab:pruning_shared}, Figure~\ref{fig:pruning}), together with two controls, random instead of Taylor importance and whole-block pruning, and the highest loss right after a pruning update, before recovery.
\begin{table*}[h]\centering\footnotesize\setlength{\tabcolsep}{2.5pt}
\begin{tabular}{llcccc}\toprule
size & setting & vanilla & \dagf{} & peak post-prune vanilla & peak post-prune \dagf{} \\\midrule
75M & unpruned finetune & 2.388 & 2.223 & - & - \\
75M & 30\% heads+channels & 2.478 & 2.298 & 2.749 & 2.578 \\
75M & 50\% heads+channels & 2.590 & 2.391 & 2.841 & 2.604 \\
75M & 70\% heads+channels & 2.829 & 2.614 & 3.199 & 2.913 \\
75M & 50\%, random importance & 3.029 & 2.935 & 4.513 & 4.729 \\
75M & 33\% whole blocks & 2.610 & 2.547 & 4.760 & 4.377 \\
75M & 50\% whole blocks & 2.832 & 3.452 & 6.232 & 7.981 \\
150M & unpruned finetune & 1.883 & 1.743 & - & - \\
150M & 30\% heads+channels & 1.957 & 1.807 & 2.199 & 2.050 \\
150M & 50\% heads+channels & 2.071 & 1.899 & 2.286 & 2.068 \\
150M & 70\% heads+channels & 2.278 & 2.091 & 2.547 & 2.317 \\
150M & 50\%, random importance & 2.280 & 2.128 & 3.240 & 3.590 \\
150M & 33\% whole blocks & 2.200 & 2.170 & 3.945 & 3.726 \\
150M & 50\% whole blocks & 2.380 & 2.245 & 5.707 & 7.929 \\
300M & unpruned finetune & 1.608 & 1.488 & - & - \\
300M & 30\% heads+channels & 1.665 & 1.541 & 1.888 & 1.762 \\
300M & 50\% heads+channels & 1.762 & 1.620 & 1.904 & 1.772 \\
300M & 70\% heads+channels & 1.967 & 1.790 & 2.087 & 2.176 \\
\bottomrule\end{tabular}
\caption{Final held-out MathInstruct NLL after 2000 finetuning steps for every pruning run of the shared-slice pairs, and the highest NLL observed immediately after a pruning update (before recovery). Whole-block pruning removes entire attention or MLP blocks; the 75M \dagf{} at half the blocks is the one setting where it loses (Section~\ref{sec:pruning}). With random unit selection both families lose far more than with Taylor importance and the \dagf{} edge shrinks at 75M.}
\label{tab:prune_all}\end{table*}

\subsection{Static substitution on all held-out sets}
\label{app:static}
Table~\ref{tab:static_all} gives the substitutions of Appendix~\ref{sec:static} for every model and held-out set. Replacing the predictor's output by its per-position mean raises the loss by at most 0.005 nats anywhere, whereas replacing the refinement's output by its mean costs 0.27 to 0.75 nats.
\begin{table*}[h]\centering\scriptsize\setlength{\tabcolsep}{3pt}
\begin{tabular}{llccccc}\toprule
model & set & intact & predictor $\to$ position mean & predictor $\to$ global mean & local $\to$ mean & both $\to$ mean \\\midrule
75M (locality arm, both) & wikitext2 & 6.164 & 6.161 & 6.162 & 6.582 & 6.595 \\
75M (locality arm, both) & mathinstruct & 4.933 & 4.928 & 4.934 & 5.267 & 5.272 \\
75M (locality arm, both) & gsm8k & 5.080 & 5.075 & 5.076 & 5.353 & 5.353 \\
75M (locality arm, both) & own & 4.009 & 4.012 & 4.038 & 4.465 & 4.502 \\
300M & wikitext2 & 3.544 & 3.547 & 3.604 & 4.251 & 4.297 \\
300M & mathinstruct & 2.565 & 2.556 & 2.600 & 3.065 & 3.065 \\
300M & gsm8k & 2.937 & 2.903 & 2.961 & 3.550 & 3.527 \\
300M & dolma21b & 2.222 & 2.223 & 2.261 & 2.920 & 2.962 \\
1B, 5B tok & wikitext2 & 3.307 & 3.310 & 3.364 & 3.996 & 4.027 \\
1B, 5B tok & mathinstruct & 2.403 & 2.405 & 2.446 & 2.940 & 2.959 \\
1B, 5B tok & gsm8k & 2.755 & 2.760 & 2.811 & 3.388 & 3.419 \\
1B, 5B tok & dolma21b & 2.076 & 2.077 & 2.117 & 2.782 & 2.812 \\
1B, 10B tok & wikitext2 & 3.219 & 3.221 & 3.281 & 3.916 & 3.950 \\
1B, 10B tok & mathinstruct & 2.321 & 2.322 & 2.366 & 2.849 & 2.868 \\
1B, 10B tok & gsm8k & 2.644 & 2.649 & 2.704 & 3.264 & 3.295 \\
1B, 10B tok & dolma21b & 2.005 & 2.005 & 2.049 & 2.758 & 2.791 \\
\bottomrule\end{tabular}
\caption{Held-out NLL when one routing channel is replaced by its mean over 128 held-out windows of the model's own corpus (`own' / `dolma21b' is that corpus's cache). The predictor's per-token output is replaceable at every size and on every set; the local refinement channel's is not.}
\label{tab:static_all}\end{table*}

\subsection{Judged chat evaluation: categories and scores}
\label{app:chat}
Tables~\ref{tab:chat_cats} and~\ref{tab:chat_scores} break down the MT-Bench comparison of Table~\ref{tab:chat_full} (80 first turns, the same judge): the win rate per category and each model's single-answer score.
\begin{table*}[h]\centering\footnotesize\setlength{\tabcolsep}{2.2pt}
\begin{tabular}{llcccccccc}\toprule
SFT & comparison & writing & roleplay & reasoning & math & coding & extraction & stem & humanities \\\midrule
Alpaca & 300M vs.\ 300M & 0.72 & 0.57 & 0.47 & 0.40 & 0.60 & 0.55 & 0.75 & 0.47 \\
Alpaca & 300M mod. vs.\ 300M & 0.80 & 0.53 & 0.42 & 0.50 & 0.50 & 0.72 & 0.55 & 0.68 \\
Alpaca & 1B@5B vs.\ 1B@5B & 0.55 & 0.60 & 0.42 & 0.45 & 0.42 & 0.62 & 0.75 & 0.85 \\
Alpaca & 1B@5B vs.\ 1B@10B & 0.42 & 0.82 & 0.50 & 0.28 & 0.53 & 0.65 & 0.53 & 0.75 \\
Alpaca & 1B@10B vs.\ 1B@10B & 0.40 & 0.75 & 0.42 & 0.33 & 0.65 & 0.65 & 0.53 & 0.75 \\
SmolTalk & 300M vs.\ 300M & 0.82 & 0.82 & 0.53 & 0.80 & 0.65 & 0.57 & 0.60 & 0.72 \\
SmolTalk & 300M mod. vs.\ 300M & 0.68 & 0.62 & 0.35 & 0.50 & 0.55 & 0.57 & 0.70 & 0.62 \\
SmolTalk & 1B@5B vs.\ 1B@5B & 0.78 & 0.62 & 0.55 & 0.53 & 0.42 & 0.57 & 0.70 & 0.53 \\
SmolTalk & 1B@5B vs.\ 1B@10B & 0.82 & 0.62 & 0.50 & 0.38 & 0.60 & 0.68 & 0.65 & 0.53 \\
SmolTalk & 1B@10B vs.\ 1B@10B & 0.80 & 0.60 & 0.78 & 0.55 & 0.50 & 0.65 & 0.57 & 0.70 \\
Alpaca & 300M s2 vs.\ 300M s2 & 0.55 & 0.47 & 0.68 & 0.60 & 0.60 & 0.60 & 0.72 & 0.62 \\
Alpaca & 300M mod. s2 vs.\ 300M s2 & 0.90 & 0.57 & 0.57 & 0.53 & 0.42 & 0.65 & 0.70 & 0.47 \\
SmolTalk & 300M s2 vs.\ 300M s2 & 0.78 & 0.68 & 0.50 & 0.68 & 0.65 & 0.47 & 0.50 & 0.88 \\
SmolTalk & 300M mod. s2 vs.\ 300M s2 & 0.53 & 0.60 & 0.50 & 0.78 & 0.62 & 0.57 & 0.50 & 0.50 \\
\bottomrule\end{tabular}
\caption{Per-category win rate of \dagf{} (10 prompts per category, so $\pm 0.15$ is noise). ``300M mod.'' is the modular variant.}
\label{tab:chat_cats}\end{table*}
\begin{table*}[h]\centering\footnotesize\setlength{\tabcolsep}{3pt}
\begin{tabular}{lccc}\toprule
model & mean score (1--10) & answer tokens & ended on end-of-turn \\\midrule
300M vanilla / Alpaca & 1.91 & 109 & 0.74 \\
300M \dagf{} / Alpaca & 2.15 & 109 & 0.72 \\
300M modular / Alpaca & 1.95 & 114 & 0.71 \\
1B vanilla 5B / Alpaca & 2.06 & 103 & 0.74 \\
1B \dagf{} 5B / Alpaca & 2.14 & 122 & 0.75 \\
1B vanilla 10B / Alpaca & 2.19 & 111 & 0.71 \\
1B \dagf{} 10B / Alpaca & 2.33 & 122 & 0.70 \\
300M vanilla / SmolTalk & 2.12 & 210 & 0.34 \\
300M \dagf{} / SmolTalk & 2.52 & 210 & 0.35 \\
300M modular / SmolTalk & 2.34 & 208 & 0.38 \\
1B vanilla 5B / SmolTalk & 2.06 & 210 & 0.34 \\
1B \dagf{} 5B / SmolTalk & 2.35 & 200 & 0.40 \\
1B vanilla 10B / SmolTalk & 2.12 & 204 & 0.38 \\
1B \dagf{} 10B / SmolTalk & 2.52 & 210 & 0.38 \\
300M vanilla seed 2 / Alpaca & 1.99 & 117 & 0.66 \\
300M \dagf{} seed 2 / Alpaca & 2.14 & 112 & 0.71 \\
300M modular seed 2 / Alpaca & 2.01 & 107 & 0.74 \\
300M vanilla seed 2 / SmolTalk & 2.16 & 211 & 0.31 \\
300M \dagf{} seed 2 / SmolTalk & 2.45 & 199 & 0.41 \\
300M modular seed 2 / SmolTalk & 2.40 & 206 & 0.38 \\
\bottomrule\end{tabular}
\caption{Single-answer judge scores (MT-Bench single-answer prompt, same judge), mean answer length and the fraction of answers that stopped on the end-of-turn token rather than the 256-token cap. Scores sit near 2 of 10 for every model at this scale; the pairwise win rate is the informative number.}
\label{tab:chat_scores}\end{table*}

\section{The predictor learns a static graph}
\label{sec:static}

What the predictor contributes is one graph per position, not per-token re-routing. Because it is a separate module, we can replace its per-token output by its mean over held-out windows (Table~\ref{tab:static_all}). At 300M and 1B the per-position mean costs at most 0.005 nats on WikiText-2 (3.544 vs 3.547; 3.219 vs 3.221 for the 1B at 10B tokens) and a single global mean 0.04 to 0.06, while the same replacement of the refinement's output costs 0.4 to 0.7 nats, more than the whole gap over the vanilla Transformer. Swapping the predictor's output between windows (activation patching; \citealp{vig2020causal,meng2022rome,zhang2024patching}) changes the loss by at most 0.007 nats from 75M to 300M (the refinement: 0.56 to 1.29).

The token input is needed to learn the graph, not to apply it. In the predictor-only arm the per-token output carries 0.03 to 0.08 nats, yet replacing it by its position mean costs only 0.01 to 0.02, so the model routes per-token variation through the refinement when it has one. A table of the same weights trained from scratch with no input recovers only 17 to 23\% of the predictor-only arm's gain over the vanilla Transformer (0.04 of 0.18 nats on the own cache, 0.06 of 0.39 on WikiText-2; Table~\ref{tab:locality}), and DenseFormer's learned static weights recover as little from 75M to 1B (at most 24\%; Table~\ref{tab:main}). The predictor is thus a parameterization that finds a good static graph, which Appendix~\ref{app:topology} analyses.

The averages and swaps above are interventions on a computation before its output exists, which ties the design to a broader aim. Post-hoc interpretability explains a computation only after it has produced its output, whereas generative interpretability asks that a model's inference pass expose checkpoints that humans can understand and that can be causally intervened on before the output is committed \citep{yang2026generative}. A predicted topology meets the second requirement by construction, since the backbone depends on it and replacing it is a well-defined intervention; whether it can also be made understandable to humans is a question this design opens.

\section{Topology analysis}
\label{app:topology}
Over 128 held-out windows of each model's own corpus we record the predictor's and the refinement's outputs. Averaged over tokens, most mixing mass sits on skip sources rather than the previous layer: the previous layer carries 17 to 61\% of the $|\alpha|$ mass for queries, 12 to 45\% for keys, only 2 to 24\% for values and 4 to 68\% for the residual (300M and 1B), and 2 to 5 sources per head exceed 5\% of the head's largest weight. Per-token spread relative to the mean is 0.05 to 0.19 for the predictor and 0.5 to 1.4 for the refinements. The graph is used densely: zeroing at inference the quarter of hyperconnection entries with the smallest mean $|\alpha|$ (both channels, per-token routing otherwise intact) costs 0.27 nats at 300M and 0.15 to 0.22 at 1B on WikiText-2, half of them costs about 1.8 nats, and removing every skip so that each layer reads only its predecessor raises the loss to 9 to 10 nats; sparsifying the topology for inference would need retraining rather than post-hoc masking.

\section{Pruning: all runs}
\label{app:prunefull}
Table~\ref{tab:pruning_full} lists every matched-corpus pruning run behind Table~\ref{tab:pruning}, and Table~\ref{tab:pruning_shared} the earlier study on the original 12B-slice pairs, which Figure~\ref{fig:pruning} plots.

\begin{table*}[h]
\centering\small
\setlength{\tabcolsep}{2pt}
\begin{tabular*}{\textwidth}{@{\extracolsep{\fill}}ll rrrr rr}
\toprule
 & & \multicolumn{4}{c}{held-out NLL $\downarrow$ at sparsity} & \multicolumn{2}{c}{NLL at 50\%, routers frozen} \\
\cmidrule(lr){3-6}\cmidrule(lr){7-8}
size & model & 0\% & 30\% & 50\% & 70\% & all routers & predictor only \\
\midrule
75M & vanilla & 2.387 & 2.475 & 2.602 & 2.815 & -- & -- \\
 & DenseFormer & 2.383 & 2.482 & 2.610 & 2.829 & 2.640 & -- \\
 & hyper-conn. & 2.306 & 2.400 & 2.526 & 2.756 & 2.517 & -- \\
 & mHC & 2.328 & 2.408 & 2.537 & 2.796 & 2.549 & -- \\
 & MUDDFormer & 2.362 & 2.461 & 2.612 & 2.889 & 2.618 & -- \\
 & \dagf{} & \textbf{2.185} & \textbf{2.258} & \textbf{2.350} & \textbf{2.669} & 2.371 & 2.365 \\
\addlinespace[2pt]
150M & vanilla & 1.854 & 1.930 & 2.038 & 2.248 & -- & -- \\
 & DenseFormer & 1.830 & 1.901 & 2.004 & 2.210 & 2.004 & -- \\
 & hyper-conn. & 1.804 & 1.878 & 1.978 & 2.190 & 1.981 & -- \\
 & mHC & 1.813 & 1.883 & 1.986 & 2.195 & 1.990 & -- \\
 & MUDDFormer & \textbf{1.692} & 1.790 & 1.908 & 2.176 & 1.905 & -- \\
 & \dagf{} & 1.724 & \textbf{1.788} & \textbf{1.882} & \textbf{2.073} & 1.894 & 1.883 \\
\addlinespace[2pt]
300M & vanilla & 1.676 & 1.722 & 1.799 & 1.975 & -- & -- \\
 & DenseFormer & 1.641 & 1.690 & 1.771 & 1.945 & 1.770 & -- \\
 & hyper-conn. & 1.604 & 1.651 & 1.727 & 1.889 & 1.730 & -- \\
 & mHC & 1.585 & 1.636 & 1.719 & 1.876 & 1.719 & -- \\
 & MUDDFormer & 1.558 & 1.611 & 1.702 & 1.901 & 1.706 & -- \\
 & \dagf{} & \textbf{1.507} & \textbf{1.549} & \textbf{1.610} & \textbf{1.756} & 1.635 & 1.613 \\
\bottomrule
\end{tabular*}

\caption{All matched-corpus pruning runs of Table~\ref{tab:pruning}: held-out MathInstruct NLL after 2000 finetuning steps of 32K tokens (units removed between steps 200 and 1400, then 600 recovery steps) at 0 to 70\% sparsity, and at 50\% with every router frozen during finetuning (DenseFormer's depth weights, the hyper-connection modules, MUDDFormer's routers, \dagf{}'s predictor and local refinement) or, for \dagf{}, only the predictor. Bold: best per size and column. }
\label{tab:pruning_full}
\end{table*}

\begin{table*}[h]
\centering\small
\setlength{\tabcolsep}{2pt}
\begin{tabular*}{\textwidth}{@{\extracolsep{\fill}}ll rrrr rr r}
\toprule
 & & \multicolumn{4}{c}{held-out NLL $\downarrow$} & \multicolumn{2}{c}{damage (\%)} & frozen \\
\cmidrule(lr){3-6}\cmidrule(lr){7-8}\cmidrule(lr){9-9}
size & model & 0\% & 30\% & 50\% & 70\% & at 50\% & at 70\% & NLL at 50\% \\
\midrule
75M & vanilla & 2.388 & 2.478 & 2.590 & 2.829 & 8.4 & 18.5 & -- \\
 & \dagf{} & \textbf{2.223} & \textbf{2.298} & \textbf{2.391} & \textbf{2.614} & 7.5 & 17.6 & 2.388 \\
\addlinespace[2pt]
150M & vanilla & 1.883 & 1.957 & 2.071 & 2.278 & 10.0 & 21.0 & -- \\
 & \dagf{} & \textbf{1.743} & \textbf{1.807} & \textbf{1.899} & \textbf{2.091} & 8.9 & 20.0 & 1.899 \\
\addlinespace[2pt]
300M & vanilla & 1.608 & 1.665 & 1.762 & 1.967 & 9.6 & 22.3 & -- \\
 & \dagf{} & \textbf{1.488} & \textbf{1.541} & \textbf{1.620} & \textbf{1.790} & 8.9 & 20.3 & -- \\
\addlinespace[2pt]
150M$^\ast$ & vanilla & 2.075 & 2.157 & 2.273 & 2.507 & 9.5 & 20.8 & -- \\
 & MUDDFormer & 1.935 & 2.037 & 2.153 & 2.366 & 11.3 & 22.3 & 2.150 \\
 & \dagf{} & \textbf{1.924} & \textbf{1.990} & \textbf{2.087} & \textbf{2.268} & 8.5 & 17.9 & 2.078 \\
\bottomrule
\end{tabular*}

\caption{The earlier pruning study on the original 12B-slice pairs (75M to 300M) and a 150M comparison with MUDDFormer on a common corpus (150M$^\ast$): held-out NLL, the performance damage relative to the same model's unpruned finetune, and the NLL at 50\% with the predictor (\dagf{}) or every router (MUDDFormer) frozen. Bold: best per block and column.}
\label{tab:pruning_shared}
\end{table*}

\section{Why reading many layers may soften pruning}
\label{app:prunewhy}
\begin{figure*}[h]
\centering
\includegraphics[width=0.82\textwidth]{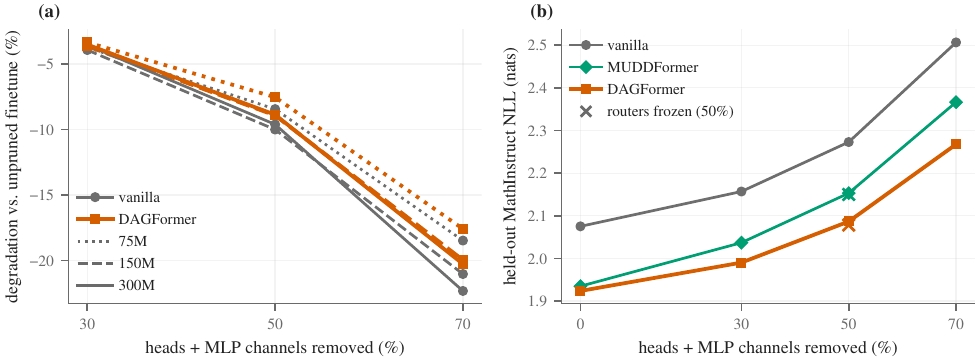}
\caption{Pruning during domain finetuning on MathInstruct. (a) Held-out NLL relative to the same model's unpruned finetune with 30/50/70\% of heads and MLP channels removed, the original 12B-slice pairs at three sizes: \dagf{} loses less at every budget, more so at higher sparsity. (b) 150M, common corpus: \dagf{} against MUDDFormer and the vanilla Transformer, equal unpruned; \dagf{} ahead by 0.05/0.07/0.10 nats at 30/50/70\%; frozen routers (crosses) change neither.}
\label{fig:pruning}
\end{figure*}
Figure~\ref{fig:pruning} plots the results of Table~\ref{tab:pruning_shared}. A head that reads several earlier layers has several routes to the information it needs, so removing one unit removes one route rather than the only one. The frozen-router result is consistent with this reading: the gain needs no re-routing, only the connections that training put in place. The random-pruning control shows its limit, since the advantage shrinks when the removed units are not the least important ones. We have not tested the redundancy hypothesis directly, for instance by counting alternative paths around removed heads.

\section{The modular variant}
\label{app:modular}
\begin{figure*}[t]
\centering
\includegraphics[width=0.72\textwidth]{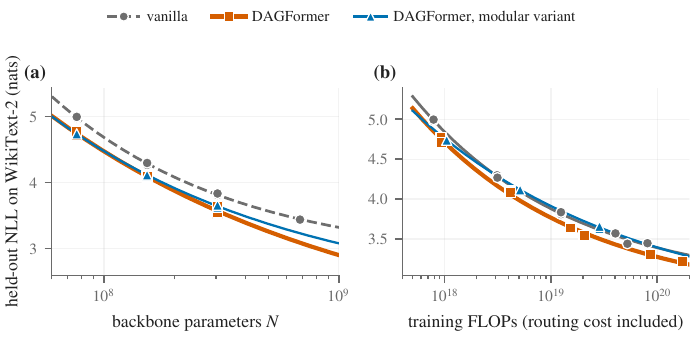}
\caption{The modular variant against \dagf{} (FourWay routing) and the vanilla OLMo-2 Transformer, as in Figure~\ref{fig:scaling}a and~b: (a) final WikiText-2 loss against backbone parameters, (b) against training FLOPs with the routing cost included. On the same corpus, modular routing is 0.03 nats behind FourWay at 75M and 150M and 0.08 behind at 300M, and per FLOP it does not repay its extra cost above 75M.}
\label{fig:modular}
\end{figure*}
Figure~\ref{fig:modular} places the modular variant in the scaling plots of Figure~\ref{fig:scaling}. In FourWay routing the sources are layer outputs, so the attention-to-MLP edge inside a layer, the block-to-everything-downstream edges and the final read-out are fixed. A \emph{modular} variant makes the sources the module outputs (embedding, then each layer's attention and MLP outputs) and adds a routed MLP input and a routed final read-out, so every intermodular edge is predicted; its routing is $2l{-}1$ sources per layer and costs 1.33$\times$ (75M) to 2.30$\times$ (300M) the vanilla Transformer's FLOPs. At 300M on the 21B corpus, FourWay beats modular on every held-out set (WikiText-2 3.544 vs 3.628, Dolma tail 2.221 vs 2.268, MathInstruct 2.565 vs 2.616, GSM8K 2.937 vs 3.093), and the modular compute multiplier is 1.15 at 75M and 0.81 to 0.93 at 150M to 300M (its extra FLOPs are not repaid). Modular routing is better in one setting: whole-block pruning scored by the mass of a module's routing column, where it beats FourWay's Taylor-pruned blocks at 150M (2.015 vs 2.044 nats at a third of the blocks removed, 2.130 vs 2.229 at half). We therefore ship FourWay and leave modular routing as an ablation.

\section{Artifacts, licenses and compute}
\label{app:artifacts}
\paragraph{Artifacts.} Every model uses the OLMo-2 architecture as implemented in Hugging Face Transformers and the OLMo-2 tokenizer \citep{olmo2} (Apache 2.0), trained from random initialization; no pretrained weights are used. Pretraining uses Dolma v1.7 (ODC-By; \citealp{soldaini2024dolma}). Instruction finetuning uses the cleaned Alpaca release (CC BY 4.0 on its dataset card; the original Alpaca data are CC BY-NC 4.0; \citealp{taori2023alpaca}), Dolly (CC BY-SA 3.0; \citealp{conover2023dolly}) and smol-smoltalk (Apache 2.0; \citealp{allal2025smollm2}). The held-out sets are WikiText-2 (CC BY-SA 3.0; \citealp{merity2016wikitext}), MathInstruct (MIT; \citealp{yue2023mammoth}), GSM8K (MIT; \citealp{cobbe2021gsm8k}) and unused Dolma files, and the evaluation tasks are those of the lm-evaluation-harness \citep{gao2024lmeval}, each under its own license. We use every artifact for non-commercial research, which all of these terms allow; all are English text. We collect no new data and rely on Dolma's own content filtering and deduplication for personal and offensive content.

\paragraph{Risks.} The models have at most 1.3B parameters and are research artifacts. Like any language model pretrained on web text, they can generate false, biased or offensive text, and they are not meant for deployment. \dagf{} changes how the layers of a Transformer are connected, not what it is trained on or for, and we see no risk specific to the architecture.

\paragraph{Compute and software.} The experiments reported here took about 3,700 GPU-hours, estimated from job records and training logs: about 2,250 on NVIDIA A100 (40GB), 510 on A40, 470 on RTX A5000, 350 on RTX A6000 and 80 on H100 GPUs. Pretraining the matched pairs and the modular variant took about 1,850 of them; the routed baselines, parameter-matched controls, learning-rate sweep and routing-source ablation about 1,200; instruction finetuning about 400; pruning about 110; and evaluation and cost benchmarks about 90. Exploratory runs not reported in the paper are not counted. We used Python 3.10, PyTorch 2.9.0, Transformers 4.57.1, Datasets 3.6.0 and lm-evaluation-harness 0.4.13 with the settings of Section~\ref{sec:expsetup}; the MT-Bench judge ran with vLLM 0.30.0, and the AlpacaEval judge through its API.

\end{document}